# Forecasting COVID-19 Case Counts Based on 2020 Ontario Data

Part of a Joint
Saint Mary's – Dalhousie – Acadia University
Research Project

funded by

Nova Scotia COVID-19 Health Research Coalition

Daniel L. Silver, Ph.D.
Rinda Digamarthi, MSc Candidate

**Acadia Institute for Data Analytics**
Acadia University

# Contents

# Glossary of Terms

- COVID-19 – Corona Virus Disease 2019
- SARS-CoV-2 – Severe Acute Respiratory Syndrome - Corona Virus -2
- IDT – Inductive Decision Trees
- ANN – Artificial Neural Networks
- RNN - Recurrent Neural Networks
- LSTM – Long Short-Term Memory
- CNN – Convolutional Neural Network
- DOY - Day of year
- DOW - Day of week
- RH – Relative Humidity
- IRH – Indoor Relative Humidity
- IAT – Indoor Air Temperature
- STL – Single Task Learning
- MTL – Multi-Task Learning
- HVAC – Heat Ventilation Air-conditioning and Cooling
- HEPA – High-Efficiency Particulate Filter
- WEKA – Waikato Environment for Knowledge Analysis
- CUDA – Compute Unified Device Architecture
- cuDNN – CUDA Deep Neural Network Library
- MAE – Mean Absolute Error
- MAPE – Mean Absolute Percentage Error
- ReLU – Rectified Linear Unit
- NSHA – Nova Scotia Health Authority
- RNS – Research Nova Scotia
- SMU – Saint Mary's University
- MDR – Meta Data Report

# Executive Summary

**Project Objective:** To develop machine learning models that are able to (1) predict the number of COVID-19 cases per day given recent weather and mobility data and (2) classify the next day as having more or less COVID-19 cases than the day before.

**Approach:** COVID-19 data from the Province of Nova Scotia was not available to the Acadia team by the time the Master's student involved (Mr. Rinda Digamarthi) need to begin his efforts so as to complete his degree in 2021. We were able to meet the objectives by using data from the four counties around Toronto, Ontario.

Data was collected and prepared into daily records containing the number of new daily COVID case counts, demographic data on those found positive with COVID, outdoor environmental conditions, typical indoor environmental conditions, and human movement based on cellular mobility and public health care restrictions.

This data was analyzed using basic descriptive statistics and linear correlation methods to determine the most important independent variables. Inductive Decision Tree (IDT) models were developed and tested to confirm variable importance and to determine the typical manner in which these variables interacted to predict daily case counts.

Models were then developed using two deep neural network approaches: Convolutional Neural Networks (CNN) and Long Short-Term Memory (LSTM) neural networks. A 5-fold chronological cross-validation approach used these methods to develop predictive models using data from March 1 – October 14, 2020, and test them on data covering October 15 - December 24, 2020, using the prior 14 days of data as input. Models were developed with and without the autoregressive Daily case counts from the prior 14 days.

**Results:** The best LSTM models forecasted **tomorrow's** daily COVID case counts with 90.7% accuracy, and the 7-day rolling average COVID case counts with 98.1% accuracy using the test set data. We also developed LSTM models to forecast **the next 7 days** of daily COVID case counts. The best models were able to predict the Daily case counts of the test set with a mean accuracy of 79.4% over all days with the lowest accuracy on the seventh day predict. Models forecasting the 7-day rolling average case counts had a mean accuracy of 83.6% on the same test set.

**Conclusions:** Most significantly, our findings point to the importance of outdoor temperature and indoor humidity for the transmission of a virus such as COVID-19. During the coldest portions of the year, when humans spend greater amounts of time indoors or in vehicles, air quality drops within buildings, most significantly Indoor Relative Humidity (IRH) levels. Moderate to high indoor temperatures coupled with low IRH (below 20%) has been known to create conditions where viral transmission via the air is more likely because: (1) water vapor ejected from an infected person's mouth can linger longer and travel further in the air because of evaporation, and (2) dry skin conditions, particularly in a recipient's airway, can make for more optimal conditions for transmission.

**Recommendations:** Within portions of buildings where the movement of persons from diverse regions is expected to meet, such as airports, seaports, train stations, bus stations, etc., the IRH should be raised to the ASHRAE recommended level of 30-50% IRH.

# 1 Introduction and Overview

This research was completed as part of a larger project entitled **"The role of environmental determinants and social mobility in viral infection transmission in Halifax"** managed by Dr. Yigit Aydede of Saint Mary's University (SMU) and funded by the Nova Scotia COVID-19 Health Research Coalition, awarded on April 29, 2020.  The researchers from SMU and Dalhousie would pursue advanced statistical approaches to understand the relationships between various independent factors and the dependent COVID-19 case count variable.  Acadia would work to develop forecasting models using the same data to predict the COVID-19 case count.

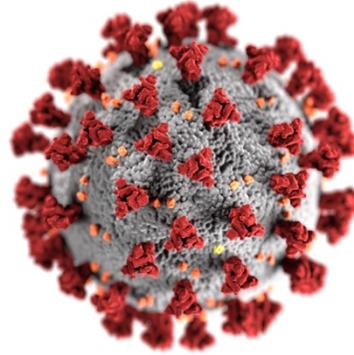

The original intention was for the joint team to work with COVID-19 data from the Province of Nova Scotia captured by the 811 Call Centre system and COVID-19 testing triage system.  Unfortunately, this data was not available three months following the award of funding, and negotiations with the province to obtain such continued for several more months.  Subsequently, the joint team agreed to proceed with data from 50 counties in Ontario.  The joint team acquired the Ontario COVID 19 data, as well as associated weather and mobility data from Apple and Google.  The SMU and Dalhousie researchers we finally able to access Province of Nova Scotia Data in the spring of 2021, but Acadia completed its work with the Ontario data.  This proved to be beneficial to forecasting as the Ontario data was a much larger and richer dataset than Nova Scotia.

**Objective:** To develop machine learning models that are able to (1) predict the number of COVID-19 cases per day given recent weather and mobility data, (2) classify the next day as having more or less COVID-19 cases than the day before.

**Approach:**  As described in Section 2, a standard data analytics project cycle was used. The work was initially intended to take place over the time frame of August 1, 2020 – March 31, 2021, with the RA working an average of 20 hours per week for 26 weeks, as he is also a full-time graduate student.  In reality, delays due to the pandemic caused the project to extend to October 2021.

# 2 Overview of Work Completed

Over the course of this project, the following people were involved:
- Danny Silver, Director and Data Scientist, Acadia Institute for Data Analytics (AIDA)
- Rinda Digamarthi, Research Assistant, Acadia Institute for Data Analytics (AIDA)
- Yigit Aydede, Overall Project Lead, Professor, Economics, Saint Mary's University
- Mutlu Yuksel, Professor, Economics, Dalhousie University (Dal)
- Ray MacNeil,  Network Manager, CLARI, Saint Mary's University (SMU)

Dr. Yigit Aydede led the overall project and assisted the Acadia team to better understand the data and the relevant statistics as well as the statistical modeling that SMU and Dal were undertaking in parallel with Acadia's forecasting effort. Dr. Danny Silver led the Acadia team and supervised Mr. Rinda Digamarthi who is using the topic for his MSc thesis. This entire data analytics project included data collection/consolidation, data preparation, predictive model development, and model evaluation. Because of the COVID-19 pandemic, most communications took place via web conferencing and telephone calls.

Phone calls and web conferences were also held with several additional parties to obtain information about:
1. The nature and status of the COVID case count data in Nova Scotia.
2. Building HVAC systems, particularly air circulation and its temperature and humidity.
3. Indoor humidity data from Ontario.

## 2.1 Work Plan

This project followed the standard data analytics process consisting of six steps: Business Understanding, Data Understanding, and Collection, Data Preparation, Model Development, Model Evaluation/Interpretation.

The work originally planned to be completed from June 1, 2020 - December 2020, was extended to December 2021 because of delays due to the COVID-19 pandemic. The RA, Mr. Rinda Digamarthi, worked an average of 20 hours per week for approximately 50 weeks while he completed the full-time graduate study.

***1. Business Understanding*** – Familiarization with the problem, terminology, and available data. This will be a shared exploratory effort with SMU. The objective is to understand the problem from a disease control perspective and to refine this into a data mining problem definition. An important aspect of this step will be ensuring the success criteria is sufficient but realistic. This project will focus on the prediction (forecasting) of the number of COVID-19 cases per day given recent weather and mobility data. Modelling techniques will be discussed and necessary changes to this Project Plan will be made during this initial step.
Milestone:     Data mining problem definition, success criteria, updated Project Plan
Resources:    D. Silver, R. Digamarthi, 1 day SMU PI
Duration:      2 weeks

***2. Data Understanding*** – Based on information obtained from the previous step, a description of the data required for modelling will be generated and agreed to by all. The source of the data will be from the following websites: https://data.ontario.ca/ - a collection of public datasets maintained by Government of Ontario, https://www.weatherstats.ca/ - weather data made available by Environment and Climate Change Canada; and
https://data.humdata.org/dataset/movement-range-maps -  mobility data from Facebook data for good website.  The SMU team will provide this data to Acadia.  The Acadia team will review the individual variables of the data coming to understand their relationship to the number of COVID-19 cases per day. It will be important for everyone to carefully understand each variable and its availability and integrity (missing values, errors).  A Meta Data Report (MDR) will be created to convey the details of syntax and semantics of the data required for the remaining steps.  Typically, this step and the next two will require two or three iterations as we work to improve the predictive models.
Milestone:     Meta Data Report
Resources:     D. Silver, R. Digamarthi, 1 day SMU PI, 3 days SMU data specialist
Duration:      4 weeks

*3. Data Preparation* – Covers all activities to construct a time series data mining data set from the original SMU data.  The Acadia team will consolidate, clean, and transform the data from the raw format to the required formats for the modelling method(s) to be used, referring to the MDR.  The daily records and their COVID-19 case counts will be prepared for training and testing the stand ML time series and deep learning methods.  This typically involves variable normalization, stationarity adjustments, and possibly dimensionality reduction.
Milestone:     Data mining dataset as per MDR
Resources:     D. Silver, R. Digamarthi, 1 day SMU PI, 1 day SMU data specialist
Duration:      6 weeks, provided we are able to access new data as needed in a timely fashion

*4. Model Development* – Deep neural networks (deep time series, and LSTM RNN) will be used to develop models given the data prepared in the previous step. This work will focus on predicting the number of new cases of COVID-19 each day, secondarily the Acadia team will look at models that classify the next day as having more or less COVID-19 cases than the day before. Sensitivity analysis will be done on the models to provide insight into the relationship between the input variables and response variable. Less accurate decision tree models may be used to provide some insight into these relationships if it seems warranted.
Milestone:     Data mining model(s)
Resources:     D. Silver, R. Digamarthi, 1 day SMU PI
Duration:      6 weeks

*5. Model Evaluation* – The model(s) must be properly tested using a cross-validation approach that will allow confidence intervals to be calculated for the models and confidence values computed for each prediction. This can be tricky for time series data as the order of the data needs to be preserved.  We initially plan to use a moving window approach that selects a chronological subset of the data and uses it to develop and test models; then the window moves along in time to provide a different collection of data. The development and testing of several models allow mean accuracy and confidence intervals to be generated. The models will be evaluated based on the success criteria established in the Business Understanding step.
Milestone:     Final predictive model(s) and results of evaluation of these models
Resources:     D. Silver, R. Digamarthi, 1 day SMU PI
Duration:      4 weeks

**6.  Iteration** – Steps 3-5 will actually be tackled in an iterative manner, with total time estimated to be that shown above.

*7. Research Report* – A research report will be created that provides details on the methods and technologies used, all prototype code that was developed, important results, and recommendations for next steps. A Report Package will be provided that contains all of this information.  This document is that report.
Deliverable:    Package of all data and code, Project Report
Resources:     D. Silver, R. Digamarthi, 2-day SMU PI
Duration:       4 weeks

# 3 Business Understanding

## 3.1 Business Problem

The main objective for the Acadia team was to develop predictive models for forecasting the spread of the COVID-19 virus and to determine the key factors affecting those predictions.  Working in tandem with the SMU and Dalhousie teams, we were to contribute to the understanding of the problem from a disease control perspective.

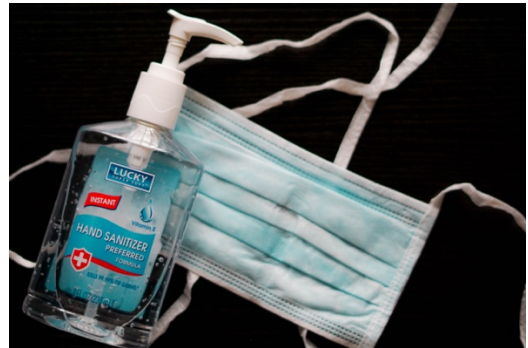

Working with the SMU and Dalhousie researchers we came to understand that there were at least five categories of variables that would need to be considered by the predictive models:
- Demographics of people (age, gender)
- Time (day of week, day of year)
- Indoor and outdoor environmental factors (temperature, relative humidity, precipitation)
- Movement of people (public health restrictions, mobility)
- Recent and current status of disease (case counts)

It would be necessary to understand the fundamental linear relationships between variables in each of the above categories with respect to COVID-19 case counts, taking into consideration various delay factors such as:
- The time for the disease to manifest symptoms in an infected person (3-14 days)
- The time before a COVID-19 sample could be taken from that person (1-7 days)
- The time required to complete the test on the sample and register it as positive (1-7)

These steps could be completed in as few as 3 days or as many as 28 days, with an estimated average of 14 days [11].

We also understood that the variables regarding the movement of people (and therefore viral spread opportunities) would be based on averages both in terms of volume of people and measurement time interval.

We also understood that the number of case counts would rise in the fall of 2020 and beyond until such time as a vaccine was discovered, which meant that we would be looking at models that would need to not just interpolate (generalize) between training data points but project to numbers of cases not previously seen.

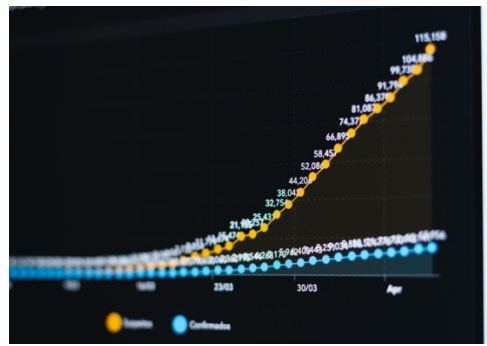

## 3.2 Data Mining Problem Definition

The technical data mining problem was refined to the following:
(1) to develop machine learning models that are able to predict the number of COVID-19 cases per day given recent demographic, time, environmental, and mobility variables, and
(2) to use those models to better understand the key factors effecting those predictions

## 3.3 Approach and Success Criteria

Considering the need for as accurate a model as possible as well as the complexity of the problem we set a success criterion of 15% MAPE (mean absolute percentage error) for the prediction of the number of COVID-19 cases per day.

We explored at several methods to model the relationship between the daily number of COVID-19 infections over space and time and the social mobility data and high-resolution weather and air quality data. Chief among these included:
- Inductive decision trees (IDT) that are good for determining linear and non-linear relationships between one or more independent variables and the dependent variables.
- Deep feed forward artificial neural networks (ANN) which could predict future days of case counts given several days' worth of independent variable values as a single input vector.
- Deeper recurrent ANNs (RNN) which could for future case counts given several days' worth of independent variable values as a sequence of daily vectors.

The first method is one of the oldest machine learning approaches that employs information theory and statistics to build a tree graph from a set of training data such that once constructed the model can accurately predict the dependent variable given a set of input variables [12]. A recursive algorithm is used to build the IDT such that the most important variable for deciding the value of the dependent variable a placed sequentially nearest the root of the tree. This approach has been used for time series prediction in the past for problems such as weather and sales prediction [13]. The IDTs would be key to determining the most import variables affecting COVID-19 case counts and therefore would be used to eliminate irrelevant variables, thereby reducing the dimensionality of the input.

The second method is an older approach that uses deep feedforward artificial neural networks and a window that moves across the spatial-temporal data capturing what is considered important independent variables over space and time to predict the dependent variable. For example, the prior week's worth of daily social mobility, weather, and air quality data along with the autoregressive terms might be used as a window to predict the next day's number of infections. This approach to time-series modeling has been used effectively for predicting the next frame in a video [1], counting people in a moving crowd [2], in medicine [3], and traffic analysis [4]. The advantage of this approach is the relevance speed and which models can be trained and tested. Recently convolutional neural networks (CNN) have been used effectively for data prepared and used in this windowing manner. The disadvantage is that the size of the window and how far it reaches back in time and spatially to capture prior variables (features) must be manually selected.

The third method is relatively new approach using deep recurrent neural networks (RNN). RNNs do not require the data to be prepared as a series of spatial-temporal windows into the past and nearby regions. The network uses recurrent connections and additional internal representation to capture the current context of the model and uses this as additional input for making predictions. Most recently, Long Short-Term Memory RNNs, or LSTMs have been used to learn long-term dependencies. They were introduced by Hochreiter & Schmidhuber (1997) and have been refined and popularized by many researchers. They have been used effectively to model complex sequences including video frames [5], spoken and written language [6], sale prediction [7], traffic flows [8], and ship movement [9]. LSTM RNNs have the advantage of learning to select the most important variables (features) from the past depending upon the most recent network context and predictions. Unfortunately, LSTMs require a lot of computing power with long training times (hours and in some cases days).

Both neural network methods require preparing the data so that it is ready to be received by the neural network machine remodels. Typically, this data engineering effort can take 50 to 70 percent of the time required by project. The project plan was refined accordingly.

# 4 Data Understanding and Collection

## 4.1 Variables Considered for Predicting the Spread of COVID-19

As mentioned in Section 3.1, there are at least five categories of variables that need to be considered for predicting changes in COVID-19 case counts. The following list provides details on each of the variables in each category that has been considered during this project. A complete Meta-Data Report can be found in Appendix B.

*Case counts:*
- Daily case count (current day = D0, yesterday = D-1, tomorrow = D+1)
    - The estimated number of cases occurring on the *Accurate_Episode_Date*. It is intended to estimate the number of new cases originating on that date, and therefore represents the number of transmitted cases.
- 7-Day average case count

- The average daily case count over the last 7 days, including today (D0). Figure 4.1 graphs these two important target variables over the 2020 study time frame.

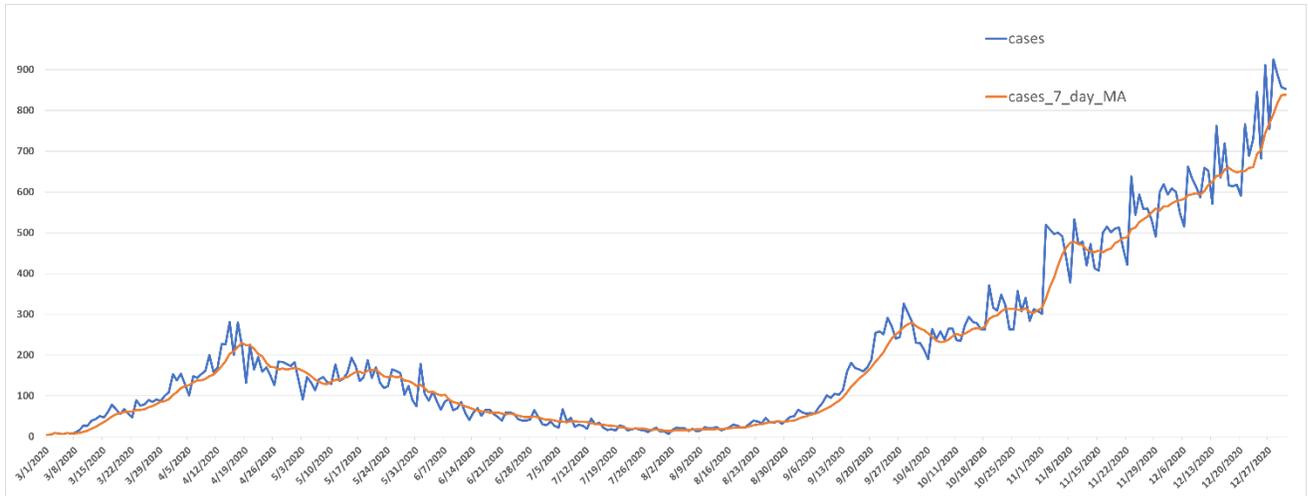

Figure 4.1(a) - The Daily case count and 7-Day average case count for 2020.

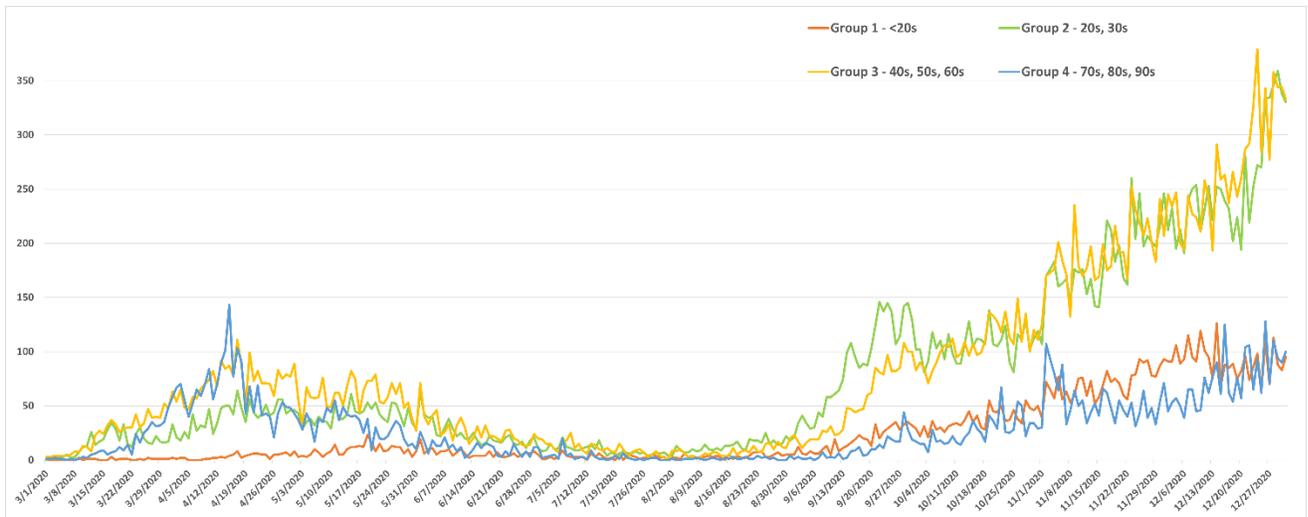

Figure 4.1(b) - The Daily case count for 2020 broken out by age.

*Demographics:*
- Age
    - The average age of the number of persons who got infected on a particular day
- Male percentage
    - The ratio of number of male cases over the Daily case count.
- Delay Mean
    - The average delay in the days between the dates when symptoms were observed and the date the person officially called the health authorities

*Transmission Type:*

- Trans-Close Contact
    - The percentage of transmissions by close contacts such as family members or friends
- Trans-Compared
    - Percentage of transmission by community spread
- Trans-Missing
    - Percentage of cases in which the information about the type of the transmission is missing from the record
- Trans-Unknown
    - Percentage of cases in which the approximate information about the type of the transmission is unknown by the patient.
- Trans-Outbreak
    - Percentage of transmission by an outbreak at a certain place. E.g.: At a sports stadium or a theatre.
- Trans-Travel
    - Percentage of transmission by air travel between the borders

*Calendar:*
- Day of year
    - The consecutive number of the day in a year
- Day of week
    - The consecutive number of the day in a week starting from Sunday

*Indoor and outdoor environmental factors:*
Maximum, Minimum, average and average hourly of the following :

- Air temperature
- Outdoor relative humidity
    - The amount of the water vapour that is present in the air in %
- Wind speed
    - The average speed at which the air moves from high to low pressure due to changes in the temperature in km/h
- Air pressure
    - The average pressure that is observed at a specific elevation and the true barometric pressure of that location
- Visibility
    - The average visibility is the maximum horizontal distance through the atmosphere that objects can be seen by the unaided eye
- Health index
    - The average air quality index value for a particular day
- Dew point
    - The value of the temperature to which air must be cooled to become saturated with the water vapour
- Precipitation

- o The amount of rain/snow received
- Snow
  - o The amount of snowfall measured in cm
- Snow on ground
  - o The amount of the snow fall which is accumulated on the ground. Measured in mm
- Wind gust
  - o A brief increase in the speed to the wind usually less than 20 seconds
- Sea Pressure
  - o The average pressure that is observed at a sea level and the true barometric pressure of that location
- Indoor relative humidity
- Maximum humidex
- Minimum windchill
  - o The effective lowering of the air temperature caused by the wind
- Heat degree days
  - o Given for each degree Celsius that the daily mean temperature departs below or above the baseline of 18 degree Celsius
- Cool degree days
  - o Given for each degree Celsius that the daily mean temperature departs below or above the baseline of 18 degree Celsius
- Grow degree days with 5C, 7C and 10C as base
- Sunrise and Sunset time
- Daylight (in hrs)
- Sunrise and Sunset forecasted
- Minimum and maximum UV forecast
- Temperature forecasts
- Solar Radiation
- Cloud cover

*Movement of people:*
- Movement_relative_to_baseline = Mobility (based on cellular phone movement) = movement of people compared to a baseline period which predates most social distancing measures.
- Proportion_users_staying_put (based on cellular phone movement) = the fraction of population that stay in place for longer time periods.
- Public health restrictions (personal bubbles, limited mobility, lock down)

## 4.2 Sources of Data Used

As stated in Section 1, unfortunately, the COVID-19 data from the Province of Nova Scotia was not available before the project needed to begin. As an alternative, the SMU team located and acquired COVID-19 case count data from 32 counties in Ontario, as well as associated weather and mobility data from Facebook social media accounts for the period from March 1 – July 31,

Lasted Revised: December 23, 2021

2020 and shared this with Acadia as a starting point for developing forecast models. This proved to be beneficial to forecasting the spread of the virus as the Ontario data had a much larger and richer dataset than that of Nova Scotia.

For the purposes of our forecast modelling efforts, we selected just four of the 32 counties making up 35% [14] of the data from Ontario; specifically, Toronto, Peel, Durham, and York counties were selected (see Figure 4.2). The total population of these counties is 5,797,924 (5.9 million) which accounts for 40% of the total population of Ontario (14,789,778).

Over the course of the next nine months the Acadia team worked to acquire values for the same variables from August 1, 2020, through to January 31, 2021, and shared this with the Dalhousie and SMU team. The COVID-19 case count data as well as the Age, Male percentage and Day of Year were extracted from the https://data.ontario.ca/ website. The weather data was extracted from https://www.weatherstats.ca/ made available by Environment and Climate Change Canada, and the mobility data was extracted from Facebook's https://data.humdata.org/dataset/movement-range-maps website.

After several weeks of search and inquiry, the Acadia team was able to locate a source of indoor relative humidity (IRH) data from the Mississauga area of Toronto in May of 2021 [16]. Thus, data was very interesting as it appeared to follow the changes in case count quite closely. We will look at this more closely in the next section.

We also compiled data on changes in Public Health Restriction from the Government of Ontario website https://covid-19.ontario.ca/ which consists of all the information on lockdowns and restrictions imposed in Ontario region from February 2020 till present.

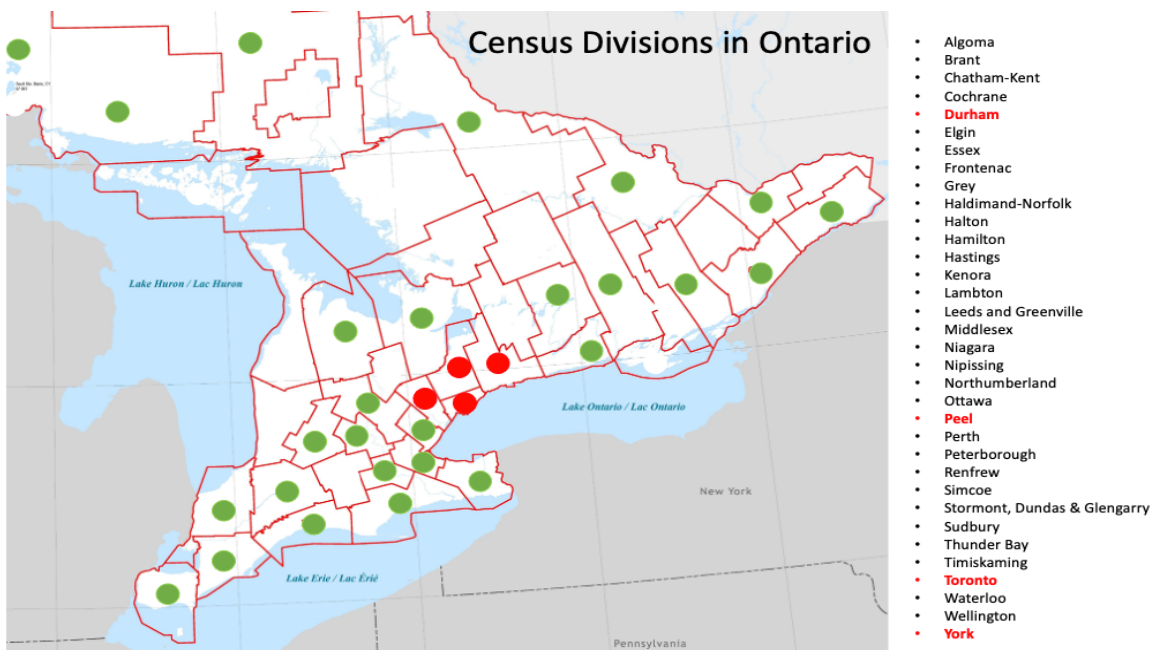

Figure 4.2- Geographic region examined in Ontario.

## 4.3 Analysis of Variables

The following summarizes an analysis of the relationships between each of the independent variables and the number of new COVID-19 case counts.

### 4.3.1 Linear Correlation

Figure 4.3 shows the correlation of each of the independent variables with the number of new case count for the current day (D0). We use this base information along with common knowledge of the relationships between variables to reduce the dimensionality of the input space. Most values with a correlation less than +/- 0.1 were removed. Furthermore, the min and max daily values for all weather variables, except *max_wind_gust*, were removed in favor of the average daily values.

Figure 4.4 and 4.5 shows the Pearson linear correlation between each variable and the number of new COVID-19 case counts as a function of lag days. Variables that show their graphs slowly increasing or decreasing for all lag values such as DOY, Restrictions, Mobility, Male percentage, and Age indicate that a shift in their calendar date into the future generally means the best correlation is with a lag of zero days. The more interesting variables in terms of lag are avg_temperature, avg_health _index, IRH, avg_visibility, DOW, precipitation, and avg_relative_humidity (outdoor humidity). We see from the graphs of the remaining variables that data from the prior 14 days has the highest correlation with the current day's case counts. Outdoor air temperature has the highest negative correlation with case counts (-0.468) when there is lag of 10 days; that is current air temperature has its highest impact on case count 10 days into the future. Similarly, IRH has the highest negative correlation (-0.301) when there is lag of 6 days, DOW (-0.114) every seventh day (Sunday), avg_visibility (-0.248) with a lag of 14 days, precipitation (-0.066) with a lag of 12 days, and avg_health_index (-0.300) with a lag >16 days).

Note that Mobility has the highest negative correlation (-0.289) with zero lag days and seems to fit in more with variables such as DOY, Restrictions, Male percentage, and Age. This is unexpected and interesting. We will present a conjecture in Section 4.6.3 that suggests that the case and effect sequence have mobility driven by Public health restrictions which is in turn driven by changes in case counts.

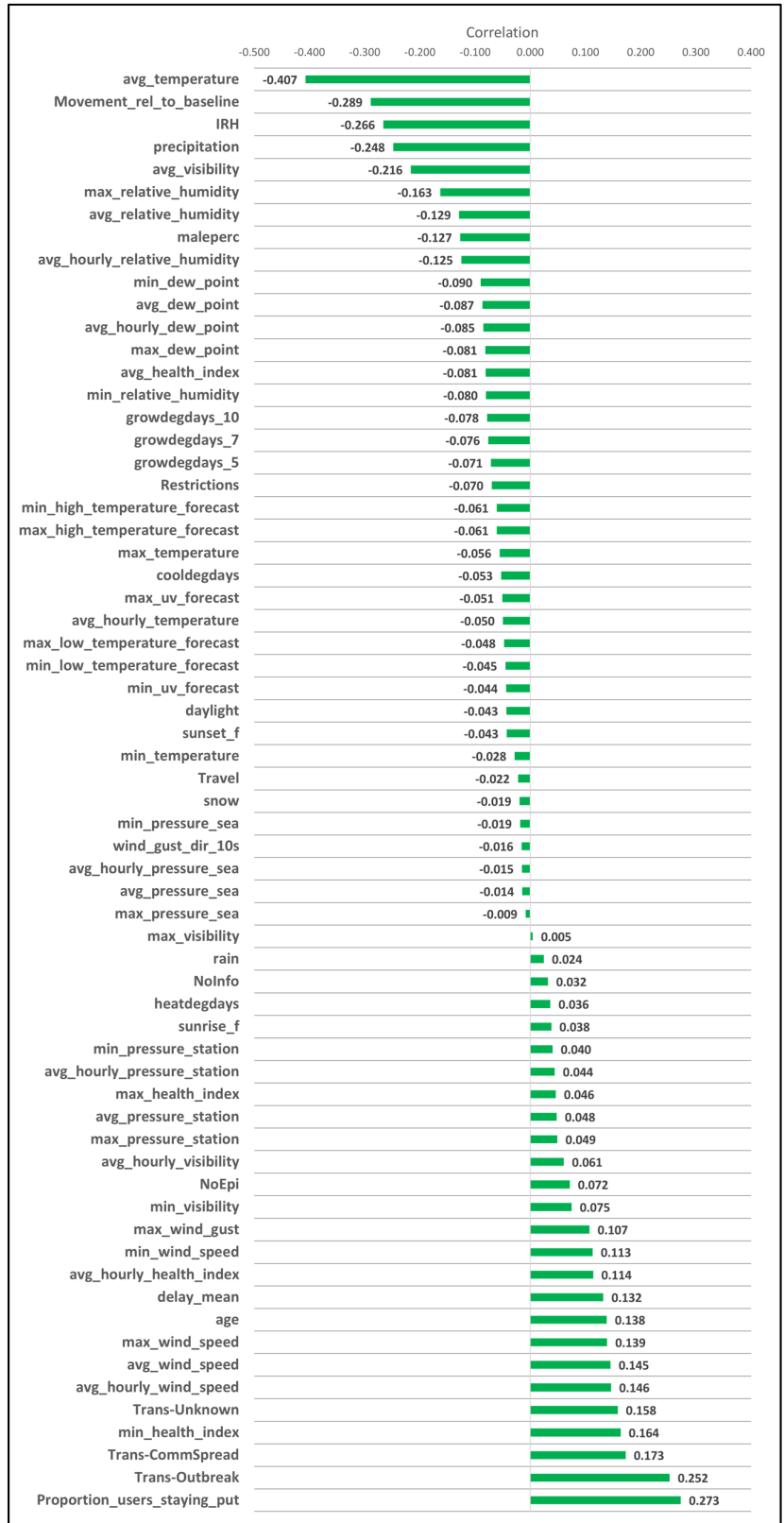

Figure 4.3 - Correlation between independent variables and current Daily case counts (D0).

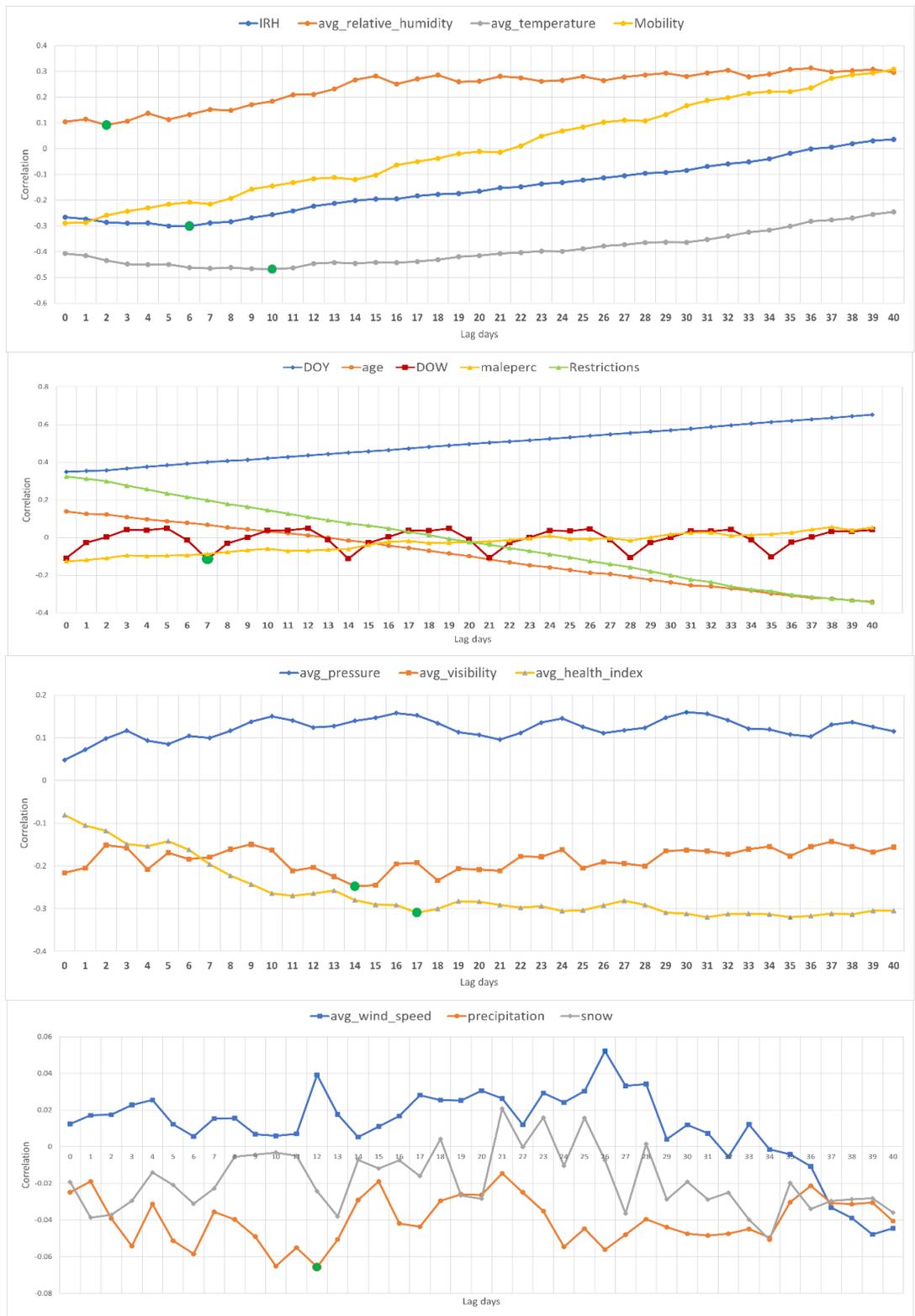

Figure 4.4 - Correlation between independent variables and case counts as a function of lag days. The green dot depicts the highest correlation value.

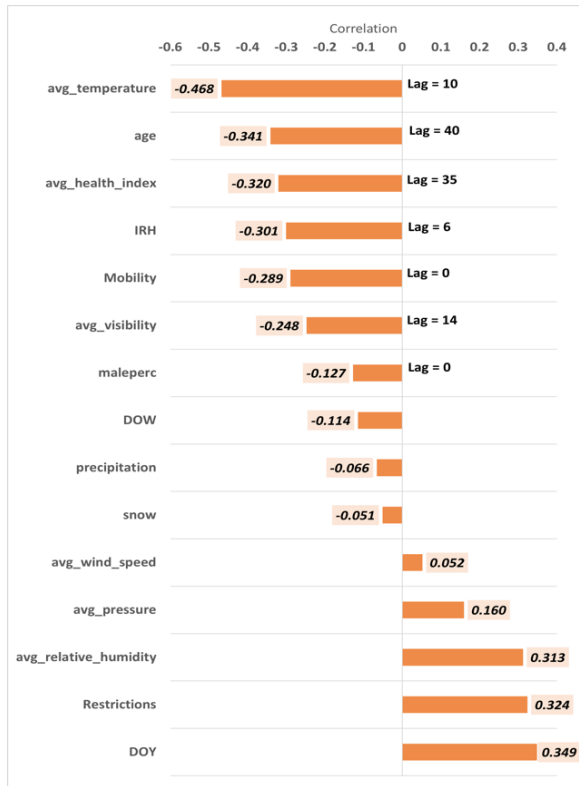

Figure 4.5: Graph of maximum correlation between independent variables and case counts.

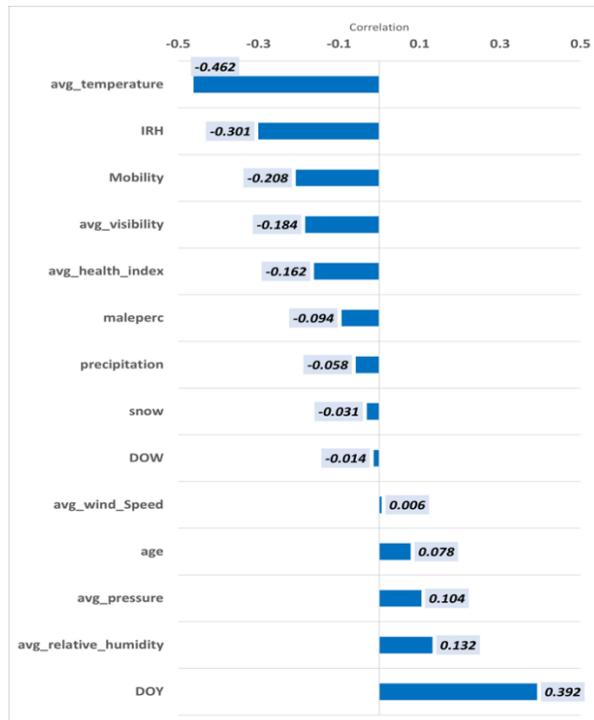

Figure 4.6 shows a bar graph of the correlation between each of major independent variables and the Daily case counts when there is a lag of 6 days.

### 4.3.2 Analysis of Daily Case Count Variables

- Daily case count - represents the number of new positive cases each day taking into consideration:
    - The time required to complete the test on the sample and registered it as positive (1-7). The date that the test is recorded is referred to as the *Test_Reported_Date*.
    - The time before a COVID-19 specimen is taken from a person (1-7 days). The date of taking the sample from the person is referred to as the *Specimen_Date*.
    - The time between the detection of symptoms and the report of a case (5-14 days). The date of reporting the case is referred to as the *Case_Reported_Date*
    - The time for the disease to manifest symptoms in an infected person following transmission.
    - The estimated date of disease transmission based on all the prior dates is referred to as the *Accurate_Episode_Date*.
    - The *delay_mean* is a function of the difference between the *Accurate_Episode_*Date and the *Case_Reported_Date*.
    - *NB:* case count could include cases recorded as few as 7 days prior to or as many as 28 days prior to the *Case_Reported_Date*, with an estimated average of 14 days prior to the *Case_Reported_Date* [11].
- 7-day average case count - the rolling average of Daily case counts over the period D-6 (6 days past) through D0 (current day)

### 4.3.3 Analysis of Demographics Variables

- Age - represents the average age of all the COVID cases on a given day. It has a significant correlation with daily COVID cases. Figure 4.1(b) shows the Daily case count by DOY broken out by Age range. This shows that in the winter and spring of 2020, those persons over 70 years contributed largely to the case counts, whereas during the fall of 2020 it was the 20- to 39-year-olds who contributed significantly. The 40-69 group contributed strongly during both periods.
- Male percentage - the percentage of the male patients on a particular day. Although it does not have a significant correlation with COVID-19, it helps to understand how male, and female are affected by it.
- During the period between March and June, the older age group were more affected by COVID, while the younger population had a greater number of COVID transmissions during the second half of the year.
- Delay_mean - the average delay in the days between when the symptoms were observed and the date when the person officially called the health authorities. It was removed because it varies widely over the study period and was thought to not be of consequence.
- The Transmission type is the suspected method of exposure to COVID-19, reported either by a patient or determined by a Public Health Unit after assessing the patient. These variables Trans-CloseContact, Trans-CommSpread, Trans-Missing, Trans-Unknown, Trans-Outbreak, Trans_Travel have a significant correlation with the Daily case counts because they naturally increase as the number of case counts increase.
- Subsequently, Trans-Close Contact, CommSpread, Missing, Unknown, Outbreak, Travel, delay_mean were not used for model development.

### 4.3.4 Analysis of Calendar Variables

- Day of year - represents numeric day of the year 1 – 365.  There is a string linear correlation to Daily case counts just because transmission of the disease increases through the year 2020.
- Day of Week - is important because transmission of diseases varies significantly from the weekday to weekend, and from day to day over the week.  The highest case counts occurred on Sundays (DoW 6) and Mondays (DoW 0) because there were more interactions between mixes of people on the weekends than during the week [15] and the recording of positive tests were higher after a weekend.  As shown in the Figure 4.7, the Mobility data shows that Saturday and Sunday (DoW 5 and 6) are the highest days of active movement in the Toronto area and Monday and Tuesday (DoW 0 and 1) tend to be the least active days.

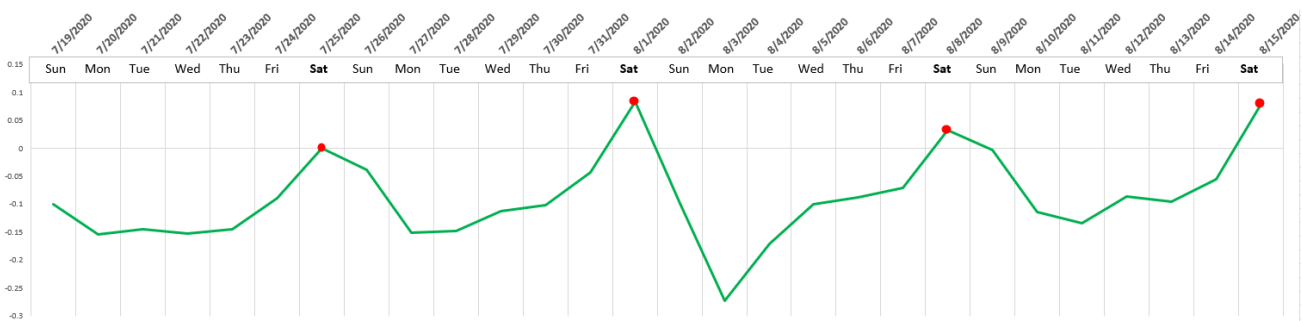

Figure 4.7 – Daily case counts versus DOW.  Peak days are normally Saturdays (DoW = 5).

### 4.3.4 Analysis of Outdoor Environmental Variables

- Outdoor air temperature - fluctuates significantly in the Toronto area; from -4.2C in March to 28.4C in August.  We see that air temperature has one of the strongest correlations with the transmission of COVID-19.  The highest negative correlation of -0.468 occurs with a lag of 10 days between the temperature and the case counts.
- Outdoor relative humidity – there appears to be only a small relationship between the spread of COVID-19 and outdoor humidity in the Toronto region of Canada.   Figure 4.9 contrasts the difference in its impact to that of indoor humidity.
- Wind speed – shows a small correlation with Daily case counts depending upon the lag
- Wind gust -  no significant correlation with Daily case counts
- Air pressure - no significant correlation with Daily case counts
- Visibility – is directly related to temperature and relative humidity; we see the highest negative correlation (-0.248) occur at a lag of 14 days.
- Health index - generally based on outdoor air quality seems to have only a minor relationship to case counts with a correlation of -0.164 at 6 days of lag
- Dew point – is calculated from outdoor air temperature and relative humidity levels, which are already available to our models
- Windchill and Humidex – are calculated from variables already available to our models, and the original records had at least 60% missing values
- Precipitation – plays a minor role, with a high negative of -0.0657 at a lag of 12 days

- Snow on ground - no significant correlation with case counts beyond that already reflected in the precipitation variable; however, it was left in as we felt that heavy snow levels could affect the ability to meet and therefore transmit the virus
- Based on their relative low correlation with the dependent Case Counts variable, the following, environment variables: heatdegdays, cooldegdays, growdegdays, Sunrise and Sunset forecast, Minimum and Maximum UV forecasts were not used for model development.

We found that all independent variables with significant correlation (avg_temperature, IRH, avg_visibility, DOW, precipitation, avg_helath_index) have a lag within 14 days, and an average of 10.5 days.

### 4.3.6 Analysis of Indoor Environmental Variables

During lower outdoor air temperatures and cold winter winds more people are working and living indoors or in vehicles for greater periods of time. This brings people into closer contact with each other and reduces their exposure to sunlight (which lowers vitamin D and melatonin levels as well as exercise and sweating). This provides a greater opportunity for viruses to spread. Building environmental conditions can contribute significantly to transmission – most significantly indoor air temperature (IAT) and indoor relative humidity (IRH).

*The ASHRAE Standard outlines that the virulence of a pathogen (harmfulness or severity) and its ability to survive depends on factors such as relative humidity, temperature, oxygen, pollutants, ozone, and ultraviolet light. Therefore, the control of indoor conditions and the use of appropriate filtration/purification methods can have a great impact on maintaining a healthy indoor environment.* [17]

The primary indoor environmental variables are:
- Density of people – controlled by Public Health Restrictions
- Quantity of fresh air – typically 10% or less in office buildings, but as much as 30% in hospitals; there are engineering standards and building manager guidelines for, but few laws
- Quality of air filtration between rooms – air re-circulation in buildings causes the same air to be inhaled by multiple individuals on the same floor or multiple floors; various levels of filtration in the HVAC system of the building are meant to reduce the spread of disease, toxic vapors, and offensive smells. Less expensive filtration methods are use in office buildings and malls, whereas more advanced and expensive HEPA filters are used in health care settings and hospitals.
- Indoor Air Temperature (IAT)– typically, kept at about 20-23.5C in summer and 23-25.5C in winter, because of variations in clothing and humidity
- Indoor Relative Humidity (IRH) – has been known for some time to have a significant impact upon the well-being of the inhabitants. The following section discusses IRH in depth because of its importance.

### *The Impact of Indoor Relative Humidity*

We consulted with two building HVAC and environmental experts in Nova Scotia, Mr. Tim Mattatal, Senior Project Manager at Stantec Consulting Ltd, Dartmouth, NS and Dr. Alain Joseph Director - Applied Research at Nova Scotia Community College, NSCC, Dartmouth, NS.

Based on discussion with them, we investigated the impact of IRH further in the literature. The following is a summary of the material that we found.

*The recommended humidity levels for a healthy and comfortable indoor environment are between 30-60% relative humidity (see Figure 4.8). The Canadian Standards Association (CSA) recommends that at a maximum, humidity levels should be between 20-70%. Relative humidity outside of these levels can cause physical discomfort, and above 70% can lead to physical damage to the building through condensation. Based on ASHRAE Standard 55 recommendations, at room temperature (21°C) RH levels should vary between about 30-75* [17].

Unfortunately, most office buildings do poorly on regulating humidity levels with IRH ranging from as high as 70% in summer to as low as 5% in winter.  This is because of costs. In summer, the cost of dehumidifying the air in the HVAC system is substantial in terms of electricity.  And in winter the cost of introducing fresh air to the HVAC system is significant in terms of heat and electricity.  Furthermore, more humid air in winter tends to increase building maintenance costs due to rust, mold, and decay.  For this reason, IRH is typically kept in the 15-25% range by building managers in winter.  But this is exactly the period when the lower outdoor temperatures force people to work and living more closely indoors or in vehicles for greater periods of time. And as shown in Figure 4.8, when IRH is below 30% the probability of transmitting disease increases significantly [19,20].

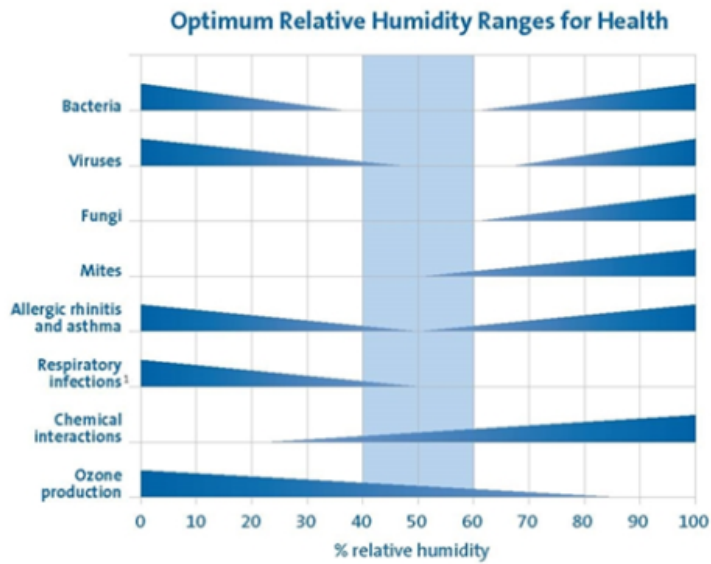

Figure 4.8 – Optimal Indoor relative humidity (IRH) levels (Ref: ASHRAE 1985) [17].

*Low relative humidity levels have been found to improve the lifetime of some viruses on hard surfaces. Airborne pathogens are also more likely to be transmitted with low levels of relative humidity. Infectious droplets from coughing and sneezing are transmitted to others by being suspended in the air. At a low humidity, water will quickly evaporate from the droplets, making them lighter and therefore easier to remain in suspension. If the relative humidity is higher, this evaporation will not happen as quickly, and the droplets are more likely to fall to the ground instead of being inhaled by another host. Therefore, to reduce the risk of infection, it is suggested that relative humidity should be maintained above 40%* [17].

*Increasing evidence points to aerosols as a main mode of transmission of COVID-19, mostly occurring indoors[1-3]. Studies suggest that environmental conditions may affect airborne transmission of respiratory viruses through different processes[4]. First, temperature (T) and relative humidity (RH) can modulate through evaporation the size distribution of exhaled aerosol particles, determining the number of particles that does not settle due to gravity and can stay suspended in the air[5, 6]. Second, they can also influence the buoyancy of the exhaled respiratory plume (which is a mixture of gases and aerosols) determining if and how it rises[7]. Third, T and RH affect the decay rate of the virus inside aerosols and droplets[6, 8]. Fourth, humidity is known to affect the immune response of the respiratory system[9]. Fifth, ambient conditions including air flow and turbulence affect transport and dispersion of the respiratory plume, and therefore the aerosol concentration at a given distance from an infected person[10]. In addition, environmental conditions may influence human behavior, such as the amount of time spent outdoors or ventilation patterns affecting the accumulation of aerosols indoors* [18].

***We found that humidity plays a prominent role in modulating the variation of COVID-19 positive cases through a negative-slope linear relationship, with an optimal lag of 9 days between the meteorological observation and the positive case report. This relationship is specific to winter months, when relative humidity predicts up to half of the variance in positive cases.*** *Our results provide a tool to anticipate local surges in COVID-19 cases after events of low humidity. More generally, they add to accumulating evidence pointing to dry air as a facilitator of COVID-19 transmission* [18].

*Our main result is that changes in daily RH anticipate changes of opposite sign in the number of COVID-19 positive cases observed 9 days later in CBA. An analysis of a subset of the data indicates that this 9-day window can be divided into a 5-day incubation period plus a 4-day testing and data processing period. Although other meteorological variables exhibited similar (weaker) relationships, our cross-validation procedures indicated that this was due to the complex set of interactions between meteorological variables. RH alone was as good in describing variations in positive cases as the whole set of variables.* ***When considering potential seasonal effects, we observed that the linear relationship between variations of RH and positive COVID-19 cases was only significant during winter months.*** *During this period, sustained monthly levels of correlation and slope were found, smoothly varying toward zero slope in the transition months. This rules out anecdotal correlations, and rather makes it likely that some behavioral pattern consistently occurring during winter, for example reduced levels of ventilation or the use of heating, is necessary for the modulation of COVID-19 trans- mission by RH.* ***Our findings provide a practical tool to predict a raise of up to 20% in positive cases following extreme low values of RH during winter months****, which could be useful for the planification of logistics in health institutions of CBA* [18].

In both the northern and southern hemispheres, the impact of humidity on viral spread has been observed particularly indoors during the colder portion of the year [18, 20]. An international study based in Europe and India [19] determined:
*The transmission routes of SARS-CoV-2 are still debated, but recent evidence strongly suggests that COVID-19 could be transmitted via air in poorly ventilated places. Some studies also suggest the higher surface stability of SARS-CoV-2 as compared to SARS-CoV-1. It is also possible that small viral particles may enter into indoor environments from the various emission sources aided by environmental factors such as relative humidity, wind speed, temperature, thus representing a type of an aerosol transmission. Here, we explore the role of relative humidity in airborne transmission of SARS-CoV-2 virus in indoor environments based on recent studies around the world. Humidity affects both the evaporation kinematics and particle growth. In dry indoor places i.e., less humidity (< 40% RH), the chances of airborne transmission of SARS-CoV-2 are higher than that of humid places (i.e., > 90% RH).*

The Acadia team was able to locate a source of IRH data from Mississauga area of Toronto [16]. This data showed a significant negative correlation between Daily case count and IRH, particularly when a lag of 6 days is considered (see Figures 4.4 and 4.5). One can see in Figure 4.9 that there is little correlation between Outdoor relative humidity and Daily case counts in the Toronto area, however there is significant correlation between Outdoor temperature and Indoor relative humidity and Daily case counts.  The lowest IRH is 10.64% in March and the highest is 59.48% in August.

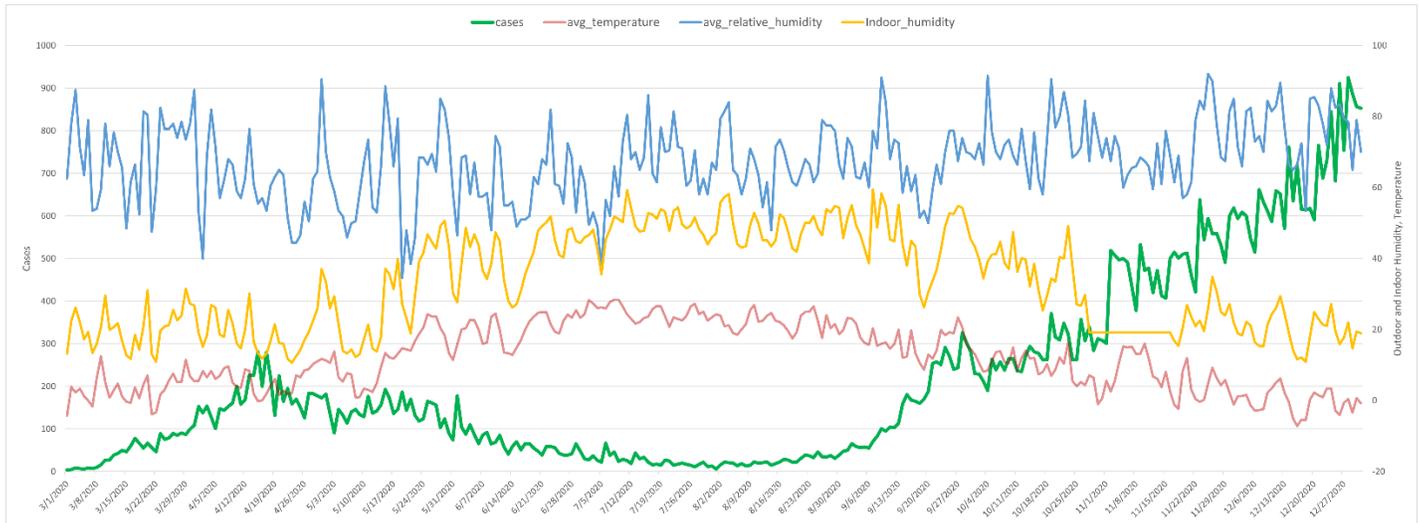

Figure 4.9 – Comparison case counts, outdoor temperature and relative humidity and indoor relative humidity.

### 4.3.7 Analysis of Movement of People Variables

- Two variables were extracted from the Facebook website https://data.humdata.org/dataset/movement-range-maps – under an initiative known as Facebook for Good have released a set of datasets to help researchers around the world to respond to COVID-19 crisis.
    - *Movement_rel_to_baseline* = Mobility - shows us movement of people compared to a baseline period which predates most social distancing measures. We use this variable as the major signal of people movement.
    - *Proportion_users_staying_put* - shows the fraction of population that stay indoors for longer time periods. It correlates strongly with Movement_relative_to_baseline so is redundant.

- Public health restrictions - We compiled data on changes in Public Health Restriction from the Government of Ontario website https://covid-19.ontario.ca/. We divided the stages of restrictions into 5 categories namely Normal, Opening 3, Opening 2, Opening 1 and Lockdown based on the dates available from the website when the restrictions have been initiated. We assigned each of these stages of Public Health Restrictions one of five levels 0-4 as shown in Table 4.1.

Figure 4.10 shows a graph of Daily case counts, Mobility, and level of Public Health Restrictions.  The graphs suggest that the cause-and-effect sequence is from Case count to changes in restrictions and finally to changes in mobility.  In fact, the linear correlation between restrictions and mobility is 0.9 with no lag and decreases with lag thereafter.   One could conclude that the rise or fall in case counts results in an increase or decrease in restrictions, followed by a decrease, or increase in mobility, respectfully. So, one can see how the restrictions worked well to reduce mobility which naturally, over time, had an impact on the transmission of the disease.

Table 4.1 Stages of Public Health Restrictions in Ontario

| Restriction | Stage Name | Indoor(max persons) | Outdoor(max persons) |
|---|---|---|---|
| 0 | Normal | - | - |
| 1 | Opening 3 | 25 | 10 |
| 2 | Opening 2 | 10 | 25 |
| 3 | Opening 1 | 5 | 10 |
| 4 | Lockdown | 5 | 0 |

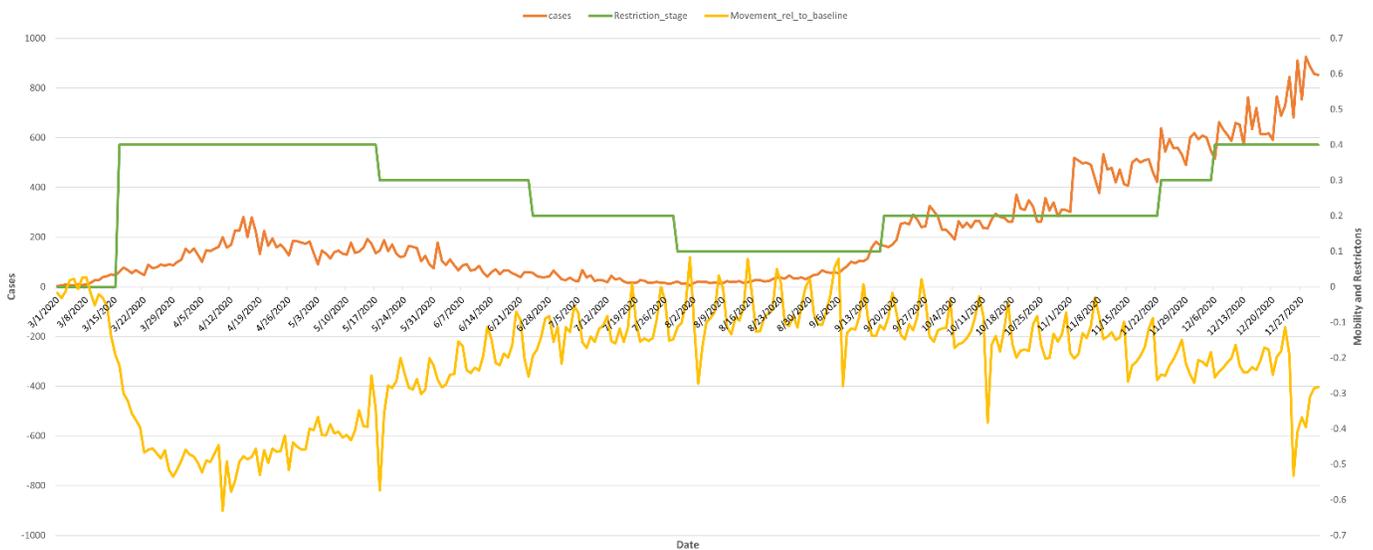

Figure 4.10 - Covid-19 Daily Case Counts versus Restrictions and Mobility.

### 4.3.8 Summary of Variables Selected for Modeling

The data suggests that in the early part of the year during the colder time frame of March and April, the transmission of COVID was highest amongst the most vulnerable demographic (older population) despite restrictions and subsequent reductions in mobility. In contrast, a reduction in restrictions along with increased mobility to normal levels in the warmer months of June through August corresponds to a reduction in Daily case count numbers to their lowest in 2020. This suggests that variables beyond human interaction are at work in COVID transmission. The fall of 2020 confirms this because as mobility slowly decreases, the number of Daily case counts soars to their highest levels.

Based on the above analysis we reduced the original set of independent variables to the list of 16 shown in Figure 4.11. These variables either had high linear correlation with Daily case counts considering a lag of up to 14 days, or they played an important combinatorial role within early inductive decision trees, as will be discussed in Section 6. We feel these variables capture the most important aspects of the demographics and movement of the population, the environments in which they were interacting, and the temporal aspects of the week or year.

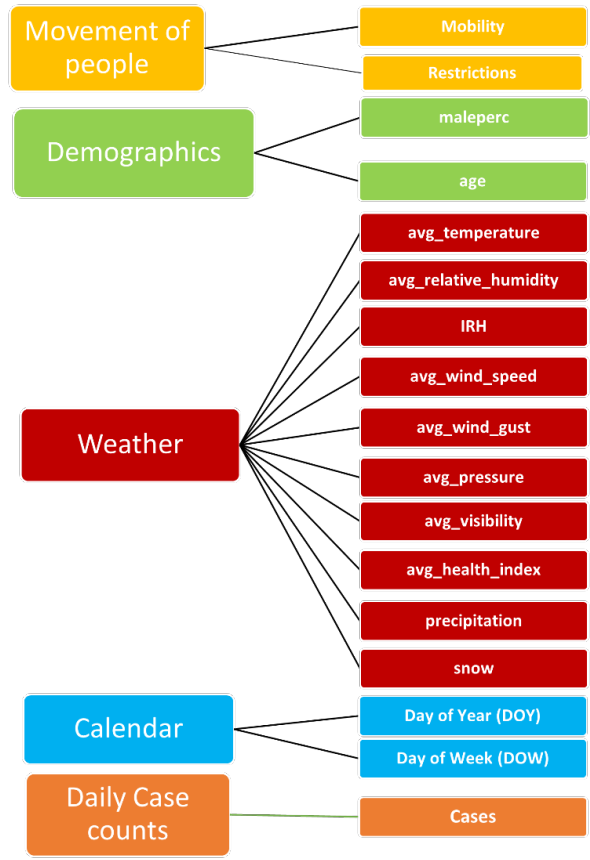

Figure 4.11 - The final list of categories and variables used to develop models.

# 5 Data Preparation

## 5.1 Data Aggregation

The data from the four sources was aggregated on a daily basis from March 1, 2020, until December 31, 2020, and then into 2021. Our studies will focus on the data from 2020 prior to the use of any vaccines in the study area. The daily case counts, mobility, indoor and outdoor environmental, and public health restrictions from Toronto, Peel, Durham, and York counties of Ontario were gathered into one spreadsheet and modified using Excel commands as well as Python code. This created a total of 306 daily records. Appendix C provides a Meta Data Report on each of these variables.

## 5.2 Data Cleaning

Of the original data consisting of case counts, calendar, weather, demographics, and mobility data, only the weather data had missing. These had to be corrected before modeling could be undertaken.

All variables which have >50% missing values were removed. This eliminated the following variables like solar_radiation, cloud_cover, snow_on_ground, max_humidex, and min_windchill. There were no variables with </=50% missing values.

## 5.3 Data Pre-processing

**Data for IDTs and CNNs:**
The initial models using IDTs were built using the past 7 days of data (D-7 to D-1, and D-0 as present) to predict one day into the future (D+1). As the IDTs do not have the ability to process sequential data, all the 7 days of past data (D-7 to D-1) has been prepared manually using the excel worksheets and with the help of python data processing techniques. Each record contains 7 days of data for each variable which is 7 days * 16 variables = 112 columns of data, plus the target case count for D+1 (or D+1 through D+7) . Eventually, we used 14 days of past data (D-14 to D-1) to predict one or more days into the future using IDTs and CNNs. This means 14 days * 16 variables = 224 columns of data.

**Data for LSTMs:**
LSTMs have the ability to accept a sequence of daily records as is without the need to preprocess it into collections of 7 or 14 days of values. Using python code, the data was structured into a 3D tensor which can be received by the TensorFlow ANN software with dimensions of samples, timesteps, and variables. Samples is the number of 7- or 14-day examples in the data, timesteps is the number of days include in each sample (7 or 14 days) and variables is the number of independent variables in each daily example.

The Meta Data Report for the final set of variables that were used to develop and test the models can be found in Appendix C.

# 6 Model Development and Evaluation

In this section we describe the hardware and software configuration used for this project. Then we describe the model development and evaluation process. First Inductive Decision Tree (IDT) models were developed to better understand which of the independent variables were most important to predicting COVID case count one day into the future (D+1) and 7 days into the future (D+1 to D+7). Following this, two types of deep learning models were developed and tested: Deep Convolutional Neural Networks (CNN) and Long Short-Term Memory Recurrent Neural Networks (LSTM RNN).

## 6.1 Hardware Configuration

During the training of deep learning and recurrent neural networks thousands of multiplication operations occur for each example in order to generate an output and then make changes to the weights of the connections of the network. A Central Processing Unit (CPU) performs these operations in a sequential manner, which is time consuming. A Graphics Processing Unit (GPU) has the ability to perform the multiplication operations in parallel, using thousands of small CPUs.

For this research we have used an inexpensive Nvidia GPU that comes with many laptops and desktop computers. The complete hardware configuration is summarized in Table 6.1.

Table 6.1: Hardware Configuration

| CPU | Intel 9th Gen Core i7 |
|---|---|
| GPU | NVIDIA GeForce GTX 1660 Ti |
| VRAM | 6 GB |
| RAM | 16 GB |

## 6.2 Software Configuration

Table 6.2 summarizes the software configuration. We used a standard, open-source data science stack which is readily accessible.

Table 6.2: Software Configurations

| Software Name | Version |
|---|---|
| Python | 3.7.4 |
| TensorFlow | 2.4 |
| Keras | 2.3.1 |
| NumPy | 1.18.1 |
| Matplotlib | 3.1.3 |
| Scikit-learn | 0.22.1 |
| CUDA | 12.1 |
| cuDNN | 8.1 |

## 6.3 Evaluation Metrics

Two metrics were used in evaluation of the IDT and ANN models on independent test sets to determine the accuracy of each model's prediction of COVID case counts. The Mean Absolute Error (MAE) and the Mean Absolute Percentage Error (MAPE).

The Mean Absolute Error measures the absolute error averaged over the entire test set. This metric can be used to compare the performance of different models on the same test set. It is given by:

$$MAE = \frac{1}{N}\sum_{i=1}^{N}|t_i - y_i|$$

Where $y_i$ is the predicted case count, $t_i$ is actual case count, and N is the total number of examples.

The Mean Absolute Percentage Error (MAPE) presents absolute error divided by the target value average over the entire test set. This metric can be used to compare the performance of different models on different test sets, so it is the most valuable. It is given by:

$$MAPE = \frac{100}{N} \sum_{i=1}^{N} ||t_i - y_i|/t_i|$$

Where necessary, repeated model builds were completed, and a hypothesis t-Test was performed to determine the significance of the performance statistic, using the p-value from the test.

# 7 Predicting COVID-19 Case Counts using IDTs

Inductive Decision Trees is one of the oldest machine learning approaches that employs information theory and statistics to build a tree graph from a set of training data such that once constructed the model can accurately predict the number of case counts given a set of input variables. A recursive algorithm is used to build the IDT such that the most important variable for deciding the value of the dependent variable a placed sequentially nearest the root of the tree. This approach has been successfully used for time series prediction in the past. IDTs can also be used to determine the most import variables affecting COVID-19 case counts and therefore can be used to eliminate irrelevant variables, thereby reducing the dimensionality of the input.

## 7.1 Experiment 1: Predicting tomorrows (D+1) case counts using 7 days of prior data

**Objective**
The goal of this experiment is to develop IDT models that accurately predict COVID case counts one day in advance (D+1) using 7 days of prior data (D-7 to D-0) and then analyse these models to determine the most important input attributes.

**Data and Methods**
This initial model used a total of 277 examples from March 1, 2020, until December 2, 2020. All variables listed in 4.3.8 were used to predict COVID case counts one day in advance (D+1). Models were developed with and without the autoregressive Daily case counts from the 7 prior days. The training set was composed of 245 examples from March 1, 2020, until October 31, and with 32 test examples from November 1, 2020. – December 2, 2020. Multiple models were developed using the WEKA Machine Learning software and tested on the independent test set, until the lowest MAPE was determined.

**Results and Discussion**

After several preliminary trials of the WEKA M5P IDT software to develop models, the best learning parameter settings were found to M=4 with Linear Regression Models at the leaves of the trees. The best IDT model using all input variables with the autoregressive variables produced an MAE = 51.32 and a MAPE = 10.90% on the independent test set. This indicates that better models can be developed using the additional variables because a simple persistence model (that uses todays case count as tomorrow's prediction) performs much worse with MAE of 88.09 cases and a MAPE of 17.64% on the same test set (based on a daily mean case count of 472 over the test set).

The best model using all input variables without the autoregressive variables produced an MAE = 61.48 and a MAPE = 13.02% on the independent test set (based on a daily mean case count of 472 over the test set). Figure 7.1 shows the graph of predicted versus actual case counts generated by this model. This is a more illuminating model because it focuses on variables other than prior case count values. Figure 7.2 shows the resulting decision tree and therefore the most important variables for forecasting Daily case counts. The most important variables were IRH, DOY, Mobility (Movement_rel_to_baseline), and avg_temperature (outdoor). The decision tree shows that an IRH less than 58% generates higher numbers of case counts. Furthermore, if the IRH is greater than 71.7%, 36 days of the training data set have relatively low case counts.

Further to this, the DOY plays an important role, determining the time of the year such as before the end of Winter 2020 (March 23), the spring and summer (March 24 – Sept 24) or the colder days after Sept 24. Within the warmer Spring and Summer period the outdoor air temperature and relative humidity play key roles in determining case counts.

A model developed without DOY but with Mobility produces an MAE = 218.25. This suggests that Mobility confounds efforts to create a general model from the available data if DOY is not also present. This was confirmed by a model with DOY but without Mobility that had an MAE = 71.18. Furthermore, a model developed with neither Mobility nor DOY, had an MAE = 80.78 on the test set. These results agree with findings of several prior studies, such as [21], that found the relationship of mobility to viral transmission to be very dependent on the time of the year, with two major phases one before and one after June 2020. As suggested, in Section 4.3.7, this may be because the cause-and-effect order is from changes in case count to changes in restrictions to finally changes in mobility.

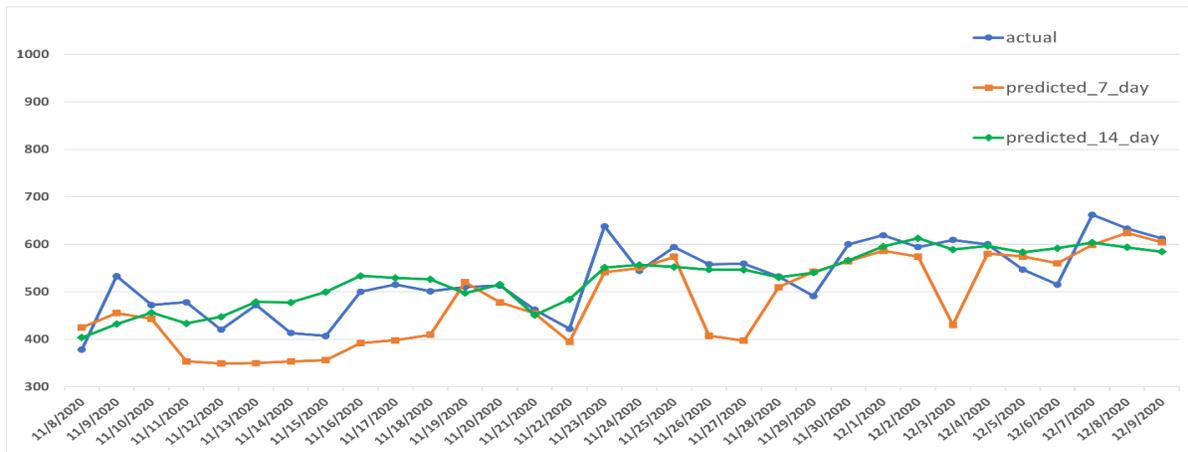

Figure 7.1 - Exp 2: Graph of 7 and 14-day IDT predictions versus actual Daily case counts.

Lasted Revised: December 23, 2021

```
D-0|IRH <= 58.45 :
|   D-6|DOY <= 257.5 :
|   |   D-6|DOY <= 83.5 :
|   |   |   D-6|Movement_rel_to_baseline <= -0.206 : LM1 (8/4.846%)
|   |   |   D-6|Movement_rel_to_baseline >  -0.206 :
|   |   |   |   D-2|Movement_rel_to_baseline <= -0.011 :
|   |   |   |   |   D-1|avg_relative_humidity <= 65.75 : LM2 (4/5.913%)
|   |   |   |   |   D-1|avg_relative_humidity >  65.75 : LM3 (7/2.216%)
|   |   |   |   D-2|Movement_rel_to_baseline >  -0.011 : LM4 (4/6.713%)
|   |   D-6|DOY >  83.5 :
|   |   |   D-6|avg_temperature <= 9.995 :
|   |   |   |   D-1|maleperc <= 0.409 :
|   |   |   |   |   D-4|avg_pressure_station <= 98.88 :
|   |   |   |   |   |   D-4|Movement_rel_to_baseline <= -0.562 : LM5 (2/0.459%)
|   |   |   |   |   |   D-4|Movement_rel_to_baseline >  -0.562 : LM6 (3/1.983%)
|   |   |   |   |   D-4|avg_pressure_station >  98.88 : LM7 (11/17.885%)
|   |   |   |   D-1|maleperc >  0.409 :
|   |   |   |   |   D-1|avg_relative_humidity <= 66.75 : LM8 (18/10.446%)
|   |   |   |   |   D-1|avg_relative_humidity >  66.75 :
|   |   |   |   |   |   D-1|avg_pressure_station <= 98.98 : LM9 (3/17.372%)
|   |   |   |   |   |   D-1|avg_pressure_station >  98.98 : LM10 (8/4.813%)
|   |   |   D-6|avg_temperature >  9.995 :
|   |   |   |   D-4|avg_pressure_station <= 100.05 :
|   |   |   |   |   D-2|IRH <= 50.55 :
|   |   |   |   |   |   D-5|maleperc <= 0.477 : LM11 (6/3.137%)
|   |   |   |   |   |   D-5|maleperc >  0.477 : LM12 (6/6.102%)
|   |   |   |   |   D-2|IRH >  50.55 :
|   |   |   |   |   |   D-0|maleperc <= 0.495 :
|   |   |   |   |   |   |   D-5|avg_relative_humidity <= 59.25 : LM13 (7/11.707%)
|   |   |   |   |   |   |   D-5|avg_relative_humidity >  59.25 : LM14 (9/4.95%)
|   |   |   |   |   |   D-0|maleperc >  0.495 :
|   |   |   |   |   |   |   D-5|avg_relative_humidity <= 67.25 : LM15 (4/10.964%)
|   |   |   |   |   |   |   D-5|avg_relative_humidity >  67.25 : LM16 (3/6.963%)
|   |   |   |   D-4|avg_pressure_station >  100.05 :
|   |   |   |   |   D-2|Movement_rel_to_baseline <= -0.274 : LM17 (3/17.991%)
|   |   |   |   |   D-2|Movement_rel_to_baseline >  -0.274 : LM18 (7/5.315%)
|   D-6|DOY >  257.5 :
|   |   D-6|avg_temperature <= 7.725 :
|   |   |   D-6|DOY <= 299.5 : LM19 (4/19.807%)
|   |   |   D-6|DOY >  299.5 : LM20 (6/9.585%)
|   |   D-6|avg_temperature >  7.725 :
|   |   |   D-6|DOY <= 285.5 : LM21 (28/22.143%)
|   |   |   D-6|DOY >  285.5 :
|   |   |   |   D-0|Movement_rel_to_baseline <= -0.143 : LM22 (7/11.045%)
|   |   |   |   D-0|Movement_rel_to_baseline >  -0.143 : LM23 (3/9.086%)
D-0|IRH >  58.45 :
|   D-5|IRH <= 71.7 :
|   |   D-2|IRH <= 63.25 :
|   |   |   D-2|avg_temperature <= 22.675 : LM24 (11/4.382%)
|   |   |   D-2|avg_temperature >  22.675 : LM25 (5/1.245%)
|   |   D-2|IRH >  63.25 :
|   |   |   D-6|DOW <= 1.5 :
|   |   |   |   D-4|Movement_rel_to_baseline <= -0.186 : LM26 (2/0.918%)
|   |   |   |   D-4|Movement_rel_to_baseline >  -0.186 :
|   |   |   |   |   D-1|avg_wind_speed <= 16.25 : LM27 (3/3.029%)
|   |   |   |   |   D-1|avg_wind_speed >  16.25 : LM28 (4/1.522%)
|   |   |   D-6|DOW >  1.5 :
|   |   |   |   D-4|age <= 41.625 : LM29 (17/4.246%)
|   |   |   |   D-4|age >  41.625 : LM30 (6/2.724%)
|   D-5|IRH >  71.7 : LM31 (36/4.484%)
```

Figure 7.2 – Exp 1: 7-day input IDT model predicting Daily case counts for D+1

## 7.2 Experiment 2: Predicting tomorrows (D+1) case counts using 14 days of prior data

**Objective**

The goal of this experiment is to develop IDT models that accurately predict COVID case counts one day in advance (D+1) using 14 days of prior data (D-14 to D-0) and then analyse these models to determine the most important input attributes. We have increased the prior number of days from 7 to 14 because we have seen from the analysis of correlations that the lag between the input variables and the COVID-19 case counts can be as much as 14 days.

**Data and Methods**

This model used a total of 277 examples from March 1, 2020, until December 2, 2020, as was the case in the first experiment, but with 14 days' worth of input. Models were developed with and without the autoregressive Daily case count variables and tested on the same independent test set, until the lowest MAPE was determined.

**Results and Discussion**

The best WEKA M5P IDT models were developed with M=4 and Linear Regression Models at the leaves of the trees. The best model using all input variables including the autoregressive variables produced an MAE = 42.65 and a MAPE = 9.30% on the independent test set (based on a daily mean case count of 472 over the test set).

The best IDT model using all input variables except the autoregressive variables produced an MAE = 53.23 and a MAPE = 11.28% on the independent test set (based on a daily mean case count of 472 over the test set). Figure 7.1 shows the graph of predicted versus actual case counts generated by this model. Clearly, the additional weeks worth of prior variables helped this new model perform better than the 7-day model.

Figure 7.3 shows a decision tree when using all but the autoregressive variables to forecast Daily case counts. The most important variables are similar to the 7-day model (DOY, Mobility (Movement_rel_to_baseline), Age, IRH, and avg_temperature (outdoor)), however, we see a heavy emphasis being placed on values from days D-13 and D-12. For example: the most important variable is the DOY from 13 days earlier (D-13) with a major break based on it being September 7 ($250^{th}$ day); this is followed by focusing on the amount of mobility that occurred 13 days earlier (D-13), if the date is prior to September 7. This makes sense since a major increase in case counts occurs in early September and the 13-day lead time would provide the opportunity for the virus to incubate and generate symptoms.

In comparison, a model developed without DOY but with Mobility produces a high MAE = 328.38. This is significantly worse than the persistence model that had an MAE of 88.09 cases and a MAPE of 17.64%. Whereas a model developed with DOY but without Mobility performed with an MAE = 79.45 which is slightly better than the persistence model. Furthermore, a model developed with neither Mobility nor DOY had an MAE = 149.43 on the test set. These results agree with the models developed using only 7 days' worth of prior data; it shows that without DOY, Mobility confounds efforts to create a general model from the available data.

```
D-13|DOY <= 250.5 :
|   D-13|Movement_rel_to_baseline <= -0.312 :
|   |   D-12|age <= 55.296 : LM1 (42/18.516%)
|   |   D-12|age >  55.296 :
|   |   |   D-1|maleperc <= 0.409 :
|   |   |   |   D-4|Movement_rel_to_baseline <= -0.562 : LM2 (2/0.415%)
|   |   |   |   D-4|Movement_rel_to_baseline >  -0.562 : LM3 (7/5.358%)
|   |   |   D-1|maleperc >  0.409 :
|   |   |   |   D-2|age <= 50.113 : LM4 (5/4.382%)
|   |   |   |   D-2|age >  50.113 : LM5 (6/7.979%)
|   D-13|Movement_rel_to_baseline >  -0.312 :
|   |   D-4|avg_temperature <= 20.57 :
|   |   |   D-13|DOY <= 237 :
|   |   |   |   D-5|Movement_rel_to_baseline <= -0.24 : LM6 (15/8.117%)
|   |   |   |   D-5|Movement_rel_to_baseline >  -0.24 : LM7 (30/9.124%)
|   |   |   D-13|DOY >  237 :
|   |   |   |   D-13|DOY <= 243.5 : LM8 (5/5.136%)
|   |   |   |   D-13|DOY >  243.5 :
|   |   |   |   |   D-12|Movement_rel_to_baseline <= -0.106 : LM9 (2/2.907%)
|   |   |   |   |   D-12|Movement_rel_to_baseline >  -0.106 : LM10 (5/3.586%)
|   |   D-4|avg_temperature >  20.57 :
|   |   |   D-7|avg_temperature <= 20.4 :
|   |   |   |   D-2|IRH_Miss <= 44.05 :
|   |   |   |   |   D-13|Movement_rel_to_baseline <= -0.255 : LM11 (3/1.794%)
|   |   |   |   |   D-13|Movement_rel_to_baseline >  -0.255 : LM12 (2/2.907%)
|   |   |   |   D-2|IRH_Miss >  44.05 :
|   |   |   |   |   D-10|avg_relative_humidity <= 78.5 :
|   |   |   |   |   |   D-11|maleperc <= 0.481 : LM13 (5/4.344%)
|   |   |   |   |   |   D-11|maleperc >  0.481 : LM14 (6/3.758%)
|   |   |   |   |   D-10|avg_relative_humidity >  78.5 : LM15 (3/2.712%)
|   |   |   D-7|avg_temperature >  20.4 :
|   |   |   |   D-10|IRH_Miss <= 48.179 : LM16 (27/8.527%)
|   |   |   |   D-10|IRH_Miss >  48.179 :
|   |   |   |   |   D-0|maleperc <= 0.592 : LM17 (17/4.676%)
|   |   |   |   |   D-0|maleperc >  0.592 : LM18 (8/2.024%)
D-13|DOY >  250.5 :
|   D-13|DOY <= 292.5 :
|   |   D-13|DOY <= 278.5 : LM19 (28/16.261%)
|   |   D-13|DOY >  278.5 :
|   |   |   D-6|Movement_rel_to_baseline <= -0.181 : LM20 (3/7.859%)
|   |   |   D-6|Movement_rel_to_baseline >  -0.181 : LM21 (11/12.558%)
|   D-13|DOY >  292.5 :
|   |   D-12|IRH_Miss <= 28.41 : LM22 (7/7.933%)
|   |   D-12|IRH_Miss >  28.41 : LM23 (6/7.856%)
```

Figure 7.3 – Exp 2: 14-day input IDT model predicting Daily case counts for D+1.

# 8 Predicting COVID-19 Case Counts using CNNs

This section records the approach and results of using deep feedforward artificial neural networks and an input window that moves across the spatial-temporal data capturing what is considered important independent variables over space and time to predict future COVID case counts. Fourteen (14) days' worth of daily social mobility, weather, and air quality data along with the autoregressive terms are used as a window to predict the next day (D+1). Recently convolutional neural networks (CNN) have been used effectively for data prepared and used in this windowing manner. The disadvantage is that the size of the window and how far it reaches back in time to capture salient variables (features) must be manually selected.  CNNs do come with the advantage of being faster to train and test than the LSTMs.

## 8.1 Experiment 3:  Predicting tomorrows (D+1) Case counts using 14 days of prior data

**Objective**
The goal of this experiment is to develop CNN models that accurately predict Daily COVID case counts for tomorrow (D+1) using a sequence of 14 days' worth of prior data.

**Data and Methods**
These models used a total of 306 examples from March 1, 2020, until December 31, 2020. All variables listed in 4.3.8 were used.  The training set was composed of 245 examples, from March 1, 2020, until October 31, 2020, validation set with 23 examples from November 1, 2020, to November 23, 2020, and with 38 test examples from November 24, 2020, till December 31, 2020.

After numerous trials of network configurations and hyperparameter settings; the best architecture was a deep convolutional neural network consisting of 2 blocks of layers; 1 convolutional block, and 1 fully connected block. The convolutional block contains one convolutional layer having a stride of 1, followed by a max pooling layer. Max-pooling is performed after each convolutional layer with a filter size of 2 and a stride of 1. Rectified Linear Unit (ReLU) is used as the activation function in each of the convolutional blocks, for the dense output layer, linear activation function is used to perform regression. This linear activation function of the output layer produces the COVID case count, which is an integer value. The Adam optimizer was used with a variable learning rate of 0.00005 and the cost function was the mean absolute error (MAE). The networks were trained for 1000 epochs, with a batch size of 1.

**Results and Discussion**
The CNN models have a MAE of 98.9 cases and a MAPE of 11.1 %. Figure 8.1 depicts the graph of predicted versus actual case counts.  These results are better than the persistence model results for the same test set with an MAPE of 14.36%.  We note that the 14-day IDT models developed had a lower MAPE of 9.3%, however this was on a different test set. We are hopefully, that the LSTM recurrent neural network approach will be superior to all of the above.

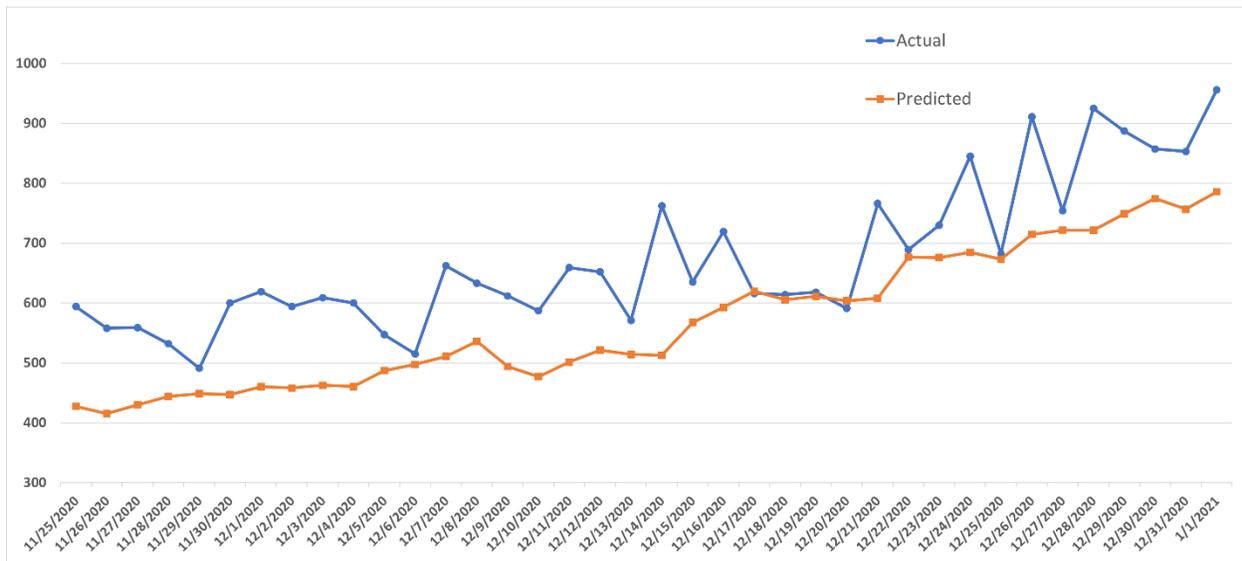

Figure 8.1 - Exp 1: CNN model predicted versus actual case counts.

# 9 LSTM RNN Models to Predict COVID-19 Case Counts

The following experiments use a deep recurrent neural network (RNN). RNNs do not require the data to be prepared as a series of spatial-temporal windows into the past and nearby regions. The network uses recurrent connections and additional internal representation to capture the current context of the model and uses this as additional input for making predictions. Specifically, we use Long Short-Term Memory (LSTM) networks. LSTM RNNs have the advantage of learning to select the most important variables from the past depending upon the most recent network context and predictions.   We compare the LSTM model performance to a baseline persistence model and to Weka M5P IDT models developed and tested on the same data.

## 9.1 Experiment 4:  Predicting tomorrows case counts using 14 days prior data with LSTMs

**Objective**
The goal of this experiment is to develop LSTM RNN models that accurately predict Daily COVID case counts for tomorrow (D+1) using a sequence of 14 days' worth of prior data.

**Data and Methods**
These models used a total of 306 examples from March 1, 2020, until December 31, 2020. The list of input variables used is shown in Section 4.3.8.  The training set was composed of 245 examples from March 1 to October 31, 2020, validation set with 23 examples from November 1 to November 23, 2020, and with 38 test examples from November 24 to December 31, 2020.

After numerous trials of network configurations and hyperparameter settings; the best architecture was a deep network consisting of 2 LSTM layers followed by one dense hidden layer of 8 nodes and then an output layer. Each LSTM layer contained 64 LSTM nodes. A Rectified Linear Unit (ReLU) is used as the activation function for each of the dense hidden layer nodes. The output node uses a linear activation function to produce the COVID case count, which is rounded to an integer value. Model development was repeated 5 times using different random initial weights and the average MAE and MAPE calculated.

**Results and Discussion**
The LSTM models have a MAE of 84.6 cases and a MAPE of 10.75%. Figure 9.1 depicts the graph of predicted versus actual case counts. These results are very similar to the deep CNN model results using 14 days' worth of input data.

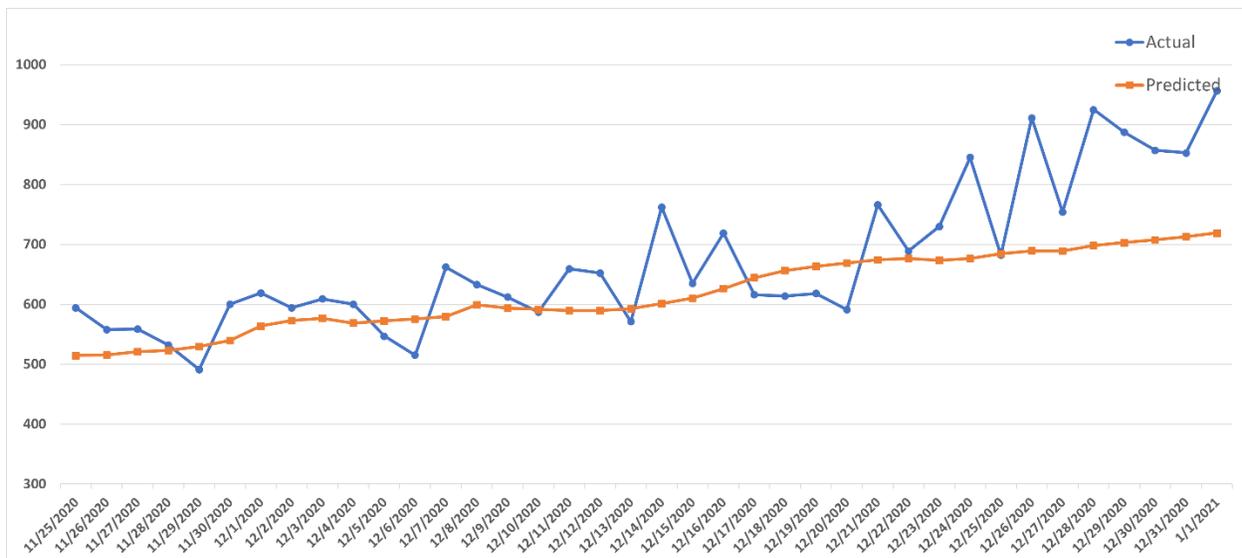

Figure 9.1 - Exp 4: Graph of predicted versus actual case counts.

## 9.2 Experiment 5: Predicting case counts for the next 7 Days using 14 days prior data

**Objective**
The goal of this experiment is to develop LSTM RNN models that accurately predict the Daily case counts for the next 7 days (D+1 through D+7) using a sequence of 14 days' worth of prior data. This is an example of a Multiple Task Learning (MTL) model.

**Data and Methods**
This model uses a total of 306 examples from March 1, 2020, until December 31, 2020, as was the case in Experiment 4. The training set is composed of 245 examples from March 1 to October 31, 2020, the validation set has 23 examples from November 1 to November 23, 2020, and the test set has 38 test examples from November 24 to December 31, 2020.

After several trials of network configurations and hyperparameter settings; the best architecture was a deep network consisting of 2 LSTM layers followed by two dense hidden layers of 128 and 64 nodes and then an output layer containing 7 nodes for the 7 days (D+1 through D+7). The first LSTM layer contains 448 LSTM nodes, and the second LSTM layer has 384. A Rectified Linear Unit (ReLU) is used as the activation function in each of the dense hidden layers and the output nodes use a linear activation function to produce the COVID case count, which we round to an integer value. Model development was repeated 5 times using different random initial weights and the average MAE and MAPE calculated.

**Results and Discussion**

The model has an average MAE of 75.12 cases and MAPE of 10.24% over all 7 days (D+1 through D+7) and an MAE of 57.10 and MAPE of 8.7% for the D+1 prediction. Figure 9.2 depicts the graph of predicted versus actual case counts for D+1. This is an excellent result, that demonstrates the value of using MTL models where related multiple related outputs are predicted by the same neural network. However, we note that the test set is slightly smaller (ending on Dec 25) because the last D+1 date is 6 days prior to the end of the test period. If we take this into consideration and test the STL LSTM model on the same portion of data, then the two models perform about the same.

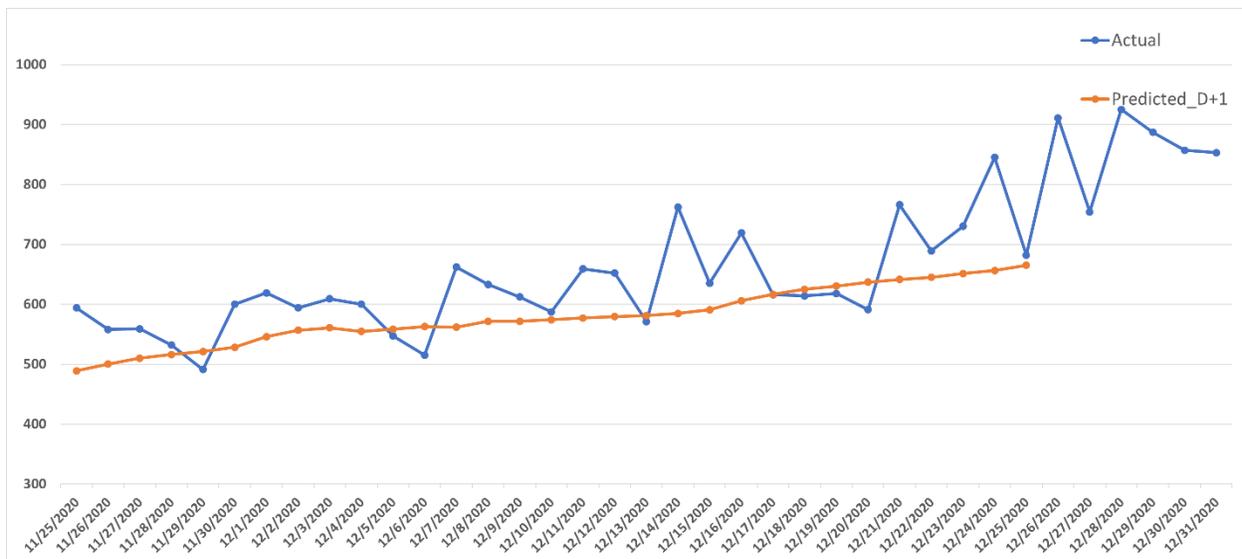

Figure 9.2 - Exp 5: Graph of predicted versus actual case counts.

## 9.3 Experiment 6: Predicting Daily case counts using k-fold chronological cross-validation

**Objective**

The goal of this experiment is to emulate the iterative development and testing of COVID forecast models over a period of several windows of data. This is what would occur in a real-world scenario. We develop LSTM STL models to predict the Daily case count for tomorrow (D+1) as accurately as possible over a period of two weeks using a sequence of 14 days' worth of prior data. We then move the window forward by two weeks. We compare the results to a baseline persistence model and to an M5P IDT model using the same data.

**Data and Methods**

These models use the same data set of 306 examples from March 1, 2020, until December 31, 2020. To develop robust models, a k-fold chronological cross-validation technique is used, where for this experiment k=5 [22]. This is meant to emulate the periodic development and use of predictive models in a real-world scenario as more data continues to arrive. The complete dataset is used to create 5 folds of training, validation, and test examples as one moves forward in time beginning from March 1, 2020. A shown in Figure 9.3, for the first fold the training and validation data is from March 1 – Oct 14, and the test set is from Oct 15 – Oct 28. For the second fold the training and validation data is from March 1 – October 28, and the test set is from Oct 29 – Nov 11. And so, on up until the last test set ending on Dec 24. The loss on the validation set is monitored using an early stopping technique to prevent overfitting. The independent test set is used to measure each model's performance and the average MAE and MAPE is reported for each output.

The same network configuration and hyperparameter settings used in Exp 4 was used for this experiment: 2 LSTM layers each of 64 nodes, followed by one dense hidden layer of 8 ReLU nodes and then the output layer. The output node uses a linear activation function to produce the COVID case count, which is rounded to an integer value. Model development was repeated 5 times using different random initial weights and the average MAE and MAPE calculated.

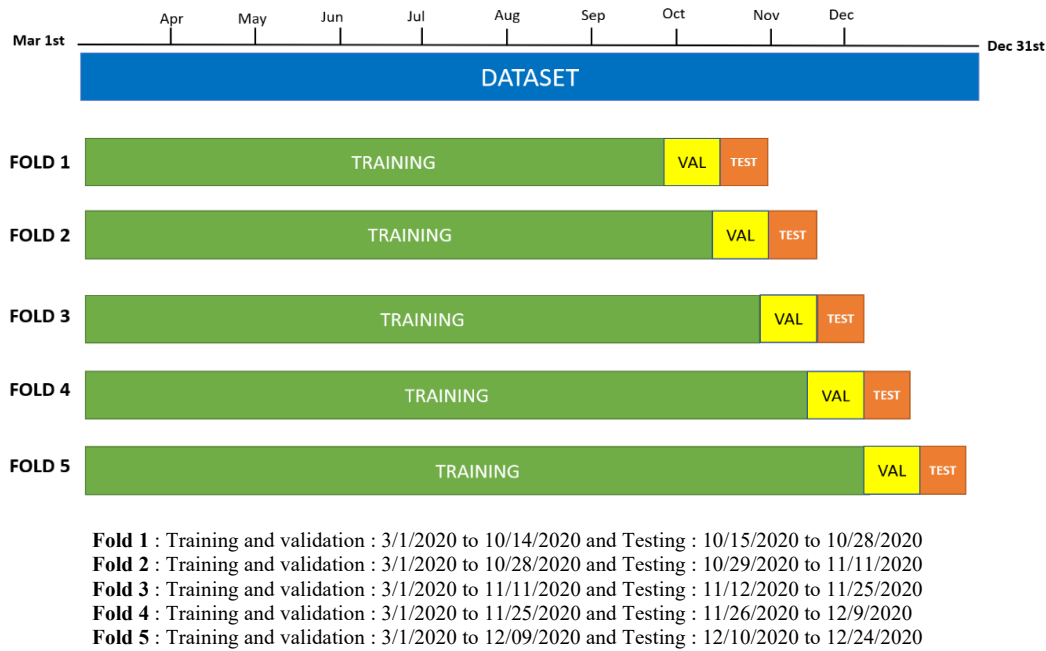

Fold 1 : Training and validation : 3/1/2020 to 10/14/2020 and Testing : 10/15/2020 to 10/28/2020
Fold 2 : Training and validation : 3/1/2020 to 10/28/2020 and Testing : 10/29/2020 to 11/11/2020
Fold 3 : Training and validation : 3/1/2020 to 11/11/2020 and Testing : 11/12/2020 to 11/25/2020
Fold 4 : Training and validation : 3/1/2020 to 11/25/2020 and Testing : 11/26/2020 to 12/9/2020
Fold 5 : Training and validation : 3/1/2020 to 12/09/2020 and Testing : 12/10/2020 to 12/24/2020

Figure 9.3 -  Exp5: The chronological 5-fold cross-validation approach used.

For comparison purposes, five M5P IDT models were developed using the same sets of training and test data. The Weka software parameters were optimized so as to develop the best possible IDT model.

**Results and Discussion**

Table 9.1 shows the performance results of the 5-fold chronological cross-validation runs for the STL models that predict the Daily case counts for D+1. Figure 9.4 shows the graphs of predicted versus actual number of cases for D+1 using these models. The models were, on average 90% accurate predicting Daile Case counts over the Oct 15 – Dec 24 period. One can see from Figure 9.4 that the model generally follows rise and fall of COVID cases over this period.

In comparison, the baseline Persistence model over the same test sets had an MAE of 74.08 cases and a MAPE of 14.02%. The M5P IDT models performed with an average MAE of 69.41 cases and MAPE of 10.48%, which is really quite good, but not to the same level as the LSTM models.

Table 9.1 – Exp 6: Performance of LSTM STL models over 5-folds

|         | MAE   | MAPE  |
|---------|-------|-------|
| Fold 1  | 24.89 | 7.82  |
| Fold 2  | 53.82 | 12.62 |
| Fold 3  | 40.70 | 7.83  |
| Fold 4  | 33.79 | 5.80  |
| Fold 5  | 64.38 | 9.75  |
| Average | 43.51 | 8.76  |
| Stdev   | 15.74 | 2.56  |

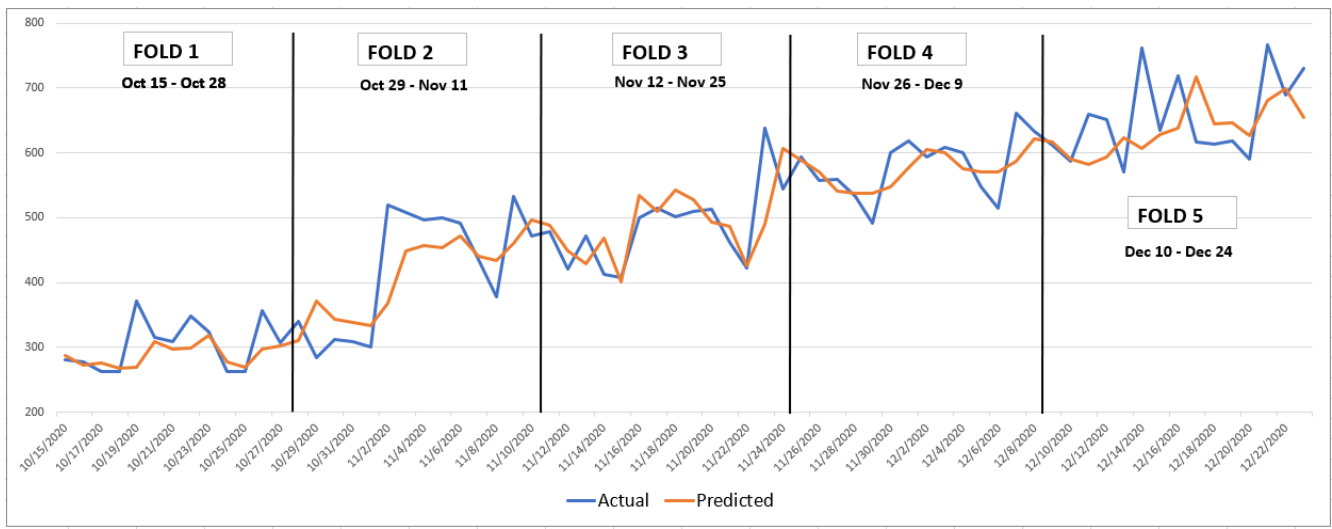

Figure 9.4 - Exp 6: LSTM STL models, predicted versus actual case counts for D+1

## 9.4 Experiment 7: Predicting 7-Day average case counts using k-fold chronological cross-validation

**Objective**

The goal of this experiment is the same as the last but for the 7-Day average case count. That is to develop LSTM STL models to predict the 7-Day average case count for tomorrow (D+1) as accurately as possible over a period of two weeks using a sequence of 14 days' worth of prior data. The window is then moved forward by two weeks. We compare the results to a baseline persistence model.

**Data and Methods**

These models use the same data set of 306 examples from March 1, 2020, until December 31, 2020. The same 5-fold chronological cross-validation technique is used as that of Exp 6; the complete dataset is used to create 5 folds of training, validation, and test examples as one move forward in time beginning from March 1, 2020. For each of the five runs, the loss on the validation set is monitored using an early stopping technique to prevent overfitting. The independent test set is used to measure each model's performance and the average MAE and MAPE is reported for each output.

The same network configuration and hyperparameter settings used in Exp 6 was used for this experiment. Model development was repeated 5 times using different random initial weights and the average MAE and MAPE calculated.

**Results and Discussion**

Table 9.2 shows the performance results of the 5-fold chronological cross-validation runs for the STL models that predict the 7-Day average case counts for D+1. Figure 9.5 shows the graphs of predicted versus actual number of cases for D+1 using these models. This method produces models that have an accuracy of between 96.21 and 99.14% over the period of Oct 15 – Dec 24, 2020. In comparison, the baseline Persistence model over the same test sets had an MAE of 7.94 cases and a MAPE of 1.74%. This surprising result is driven by the smooth-flowing change in the 7-Day average as compared to the more stochastic nature of the Daily case count (see Figure 4.1(a)). Naturally, a simple persistent model will not work well for predicting beyond the next time step, as today's case count value increasingly is different from the actual case count at D+3, D+4, or D+7.

Table 9.2 – Exp 7: Performance of LSTM STL models over 5-folds

|         | MAE   | MAPE |
|---------|-------|------|
| **Fold 1** | 4.07  | 1.40 |
| **Fold 2** | 14.48 | 3.79 |
| **Fold 3** | 9.17  | 1.68 |
| **Fold 4** | 9.09  | 1.60 |
| **Fold 5** | 5.52  | 0.86 |
| **Average** | 8.47 | 1.87 |
| **Stdev** | 4.03  | 1.12 |

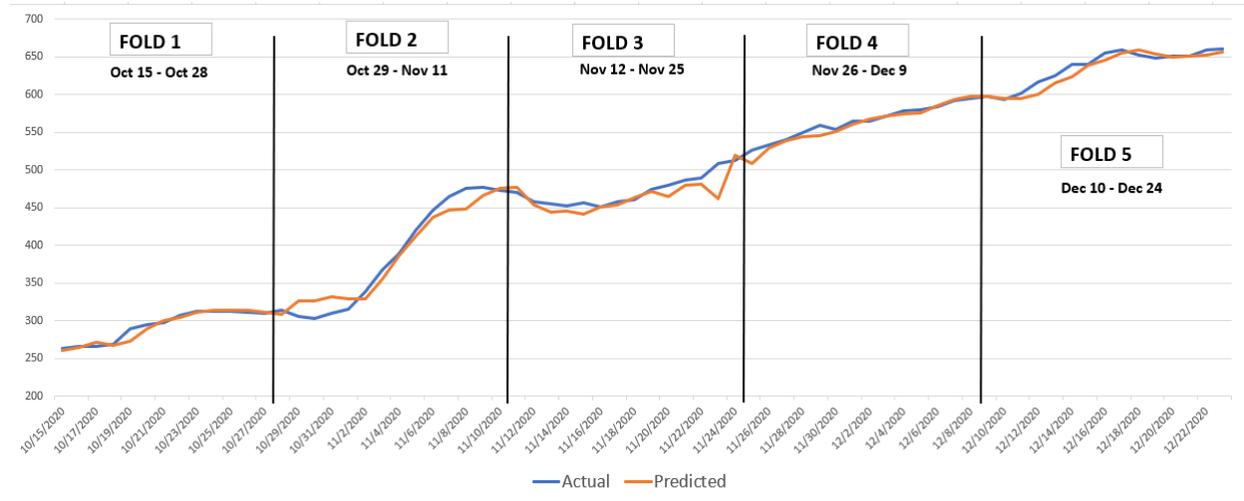

Figure 9.5 -  Exp 7: LSTM STL models, predicted versus 7-Day average case counts for D+1

# 10 Summary and Recommendations

## 10.1 Summary and Conclusions

An analysis of the data over 2020 suggests that in the early part of the year during the colder time frame of March and April, the transmission of COVID was highest amongst the most vulnerable demographic (older population) despite restrictions and subsequent reductions in mobility (see Figure 4.1(b)).  In contrast, a reduction in restrictions along with increased mobility to normal levels in the warmer months of June through August corresponds to a reduction in Daily case count numbers to their lowest in 2020 (See Figure 4.10).  This suggests that variables beyond human interaction are at work in COVID transmission.  The fall of 2020 confirms this because as mobility slowly decreases, the number of Daily case counts soars to their highest levels largely in the 20-39-year-olds (See Figure 4.1(b)).

Guided by the above analysis we reduced the original set of independent variables to the list of 16 shown in Figure 4.11.  These variables either had high linear correlation with Daily case counts considering a lag of up to 14 days, or they played an important combinatorial role within early inductive decision trees (IDT), as discussed in Section 6.  These variables capture the most important aspects of the demographics and movement of the population, the environments in which they were interacting, and the temporal aspects of the week or year.

Predictive models were developed using two deep neural network approaches:  Convolutional Neural Networks (CNN) and Long Short-Term Memory (LSTM) neural networks.  A 5-fold chronological cross-validation approach employed these methods to develop predictive models using data from March 1 – October 14, 2020 and test them on data covering October 15 - December 24, 2020.  The prior 14 days of data as input is used to train and test the models.  The best models for predicting tomorrow's (D+1) Daily case count and 7-Day average case count were based on the LSTM deep neural network architecture. Each model used 14 days' worth of prior data (including the Daily case count autoregressive values) as input and predicted tomorrow's case count, day D+1. The best Daily case count model had an MAE of 44.70 counts and an MAPE of 9.23% with a 95% CI of 2.03%. The best 7-Day average case count model had an MAE of 8.47 counts and an MAPE of 1.87% with a 95% CI of 0.98%.

The models show that the Day of Year (DOY), Mobility (Movement_rel_to_baseline) and Age naturally played important roles in the models but in different ways depending upon if it is before or after June 2020. Figure 4.10 shows a graph of Daily case counts, Mobility, and level of Public Health Restrictions. The graphs suggest that the cause-and-effect sequence is from Case count to changes in restrictions and finally to changes in mobility. In fact, the linear correlation between restrictions and mobility is 0.9 with no lag and decreases with lag thereafter. One could conclude that the rise or fall in case counts results in an increase or decrease in restrictions, followed by a decrease, or increase in mobility, respectfully. One can see from Figure 4.10 how the restrictions worked well to reduce mobility which naturally, over time, had an impact on the transmission of the disease.

Figure 4.1(b) shows that during the first wave of 2020, those persons over 70 years of age contributed largely to the case counts, whereas during the second wave it was the 20- to 39-year-olds who contributed significantly. The 40-69 group contributed strongly during both periods. Therefore, we believe that the Public Health Restrictions and environment driven by the seasons played a significant role in determining which age groups were affected the most. There are lessons to be learned from this and they are reflected in the recommendations below.

We found the most significant and interesting environmental variables affecting the models were Outdoor average air temperature (avg_temperature) and Indoor Relative Humidity (IRH) as shown in Figure 4.9. Related studies from all over the world have shown that outdoor air temperature plays a significant role in the transmission of COVID-19 virus. During the coldest portions of the year when humans spend greater amounts of time indoors or in vehicles, air quality drops within buildings, most significantly Indoor Relative Humidity (IRH) levels. Moderate to high indoor temperatures coupled with low IRH (below 20%) has been known to create conditions where viral transmission via the air is more likely because: (1) water vapor ejected from an infected person's mouth can linger longer and travel further in the air because of evaporation, and (2) dry skin conditions, particularly in a recipient's airway, can make for more optimal conditions for transmission.

## 10.2 Recommendations

We recommend the following actions based on our analysis of the data and the predictive models developed from the data of the Toronto area:

- The importance that indoor environment plays on the transmission of disease during colder portions of the year needs to be better communicated. In portions of buildings where persons from different regions or "bubbles" are expected to meet, such as airports, seaports, train stations, and bus stations the IRH should be raised to the ASHRAE recommended level of

- 30-50% IRH.  The cost of doing this is recognized, and so difficult decisions will need to be taken.
- Early Public Health Restrictions should focus on the most vulnerable portion of the population first, typically seniors.  In the case if COVID-19-like diseases, restrictions, and changes in operating procedures specific to senior facilities and nursing homes should be planned for, tested, and redesigned as needed so as to reduce the chances of the disease making its way to the most vulnerable at the start of the pandemic.
- Future modeling should focus on modeling the general trend in disease transmission provided by the 7-Day moving average case count versus the Daily case count.  This is largely because of the noise introduced by estimates of when a person contracted the disease driven by variations in the testing and reporting sequence during the pandemic.
- Multiple task learning models developed to predict Daily case counts or transmission rates for several days into the future (e.g., D+1 through D+7) seem to have the advantage over standard single task models in terms of predictive performance.  This needs to be investigated further.

## Acknowledgements

We would like to thank the following people and organizations:
- Dr. Yigit Aydede of  Saint Mary's University for his leadership and statistical guidance as PI on this project.
- Mr. Ray MacNeil from CLARI for bringing the team together and for his counsel and meeting administration over the course of the project.
- Dr. Mutlu Yuksel of Dalhousie University for his feedback and collegial interactions during the project.
- The NS COVID 19 Health Research Coalition for the financial support for this project.
- Acadia University for their administrative and in-kind support.

# Appendices

## A. Research Ethics Board Approval

**From:** Stephen Maitzen <stephen.maitzen@acadiau.ca>
**Date:** Monday, September 14, 2020 at 11:03 AM
**To:** Daniel Silver <danny.silver@acadiau.ca>
**Subject:** RE: REB Application

Dear Dr. Silver,

Thank you for contacting me. Because your research involves only the analysis of anonymized data that were originally collected for a different purpose and that cannot be linked to individuals, your project is exempt from ethics review under TCPS2 Article 2.4. You do not need to submit a research ethics application.

Best wishes for a successful project.

****************************
Stephen Maitzen, PhD
W. G. Clark Professor of Philosophy
Head, Department of Philosophy
Chair, Research Ethics Board
Acadia University

**From:** Danny Silver <danny.silver@acadiau.ca>
**Sent:** Monday, September 14, 2020 10:11 AM
**To:** Stephen Maitzen <stephen.maitzen@acadiau.ca>
**Subject:** REB Application

Hello  Dr. Maitzen (Stephen) . Please find attached an application for Ethical Review of Research.
This is a joint project with Saint Mary's and Dalhousie where we are using data concerning COVID-19 collected and aggregated by health authorities in Ontario and Nova Scotia.   We will be using counts of persons who have approached 811 with symptoms by region by date or have been found positive with COVID based on a test by region by date.  No personal information is included in the data from the provinces which is publicly available in Ontario and will soon be in NS (we hope).

I have attached the review response from SMU for your information.
They did not require a full review.  I can send you their REB app, if you wish.

============================
**Daniel L. Silver**
Professor, Jodrey School of Computer Science
Director, Acadia Institute for Data Analytics
Acadia University,
Office 314, Carnegie Hall,
Wolfville, Nova Scotia Canada B4P 2R6

t. (902) 585-1413
f. (902) 585-1067

**Research Ethics Board**   Acadia University Box 181
Wolfville, Nova Scotia
Canada  B4P 2R6
http://reb.acadiau.ca

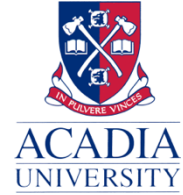

# Application for Ethical Review of Research Involving Humans

Complete this form electronically and submit it, along with your **Application Package**, by **email attachment** to smaitzen@acadiau.ca. Please attach it as a single **Microsoft Word** or **PDF** file. No digital signature is required on your documents.

The Research Ethics Board (REB) strongly encourages you to consult the *Tri-Council Policy Statement, Second Edition* (TCPS2), when preparing your application. TCPS2 can be found at this link. **Incomplete forms will be rejected**.

| | |
|---|---|
| Name of Principal Investigator: | Daniel L. Silver |
| Faculty, Staff, Graduate student, Undergraduate student? | Faculty |
| Department, School, or Program: | Jodrey School of Computer Science |
| Telephone number: | cell:  902-679-9315 / office: (902) 585-1413 |
| Email address: | danny.silver@acadiau.ca |
| Supervisor (if you are a student): | |
| Supervisor's email address: | |
| Title of your project: | The role of environmental determinants and social mobility in viral infection transmission in Ontario and Nova Scotia. |

Type of project (e.g., Honours or Master's thesis; externally funded project; part of a research program):

| |
|---|
| Masters thesis, funded by the Nova Scotia COVID-19 Health Research Coalition, part of an SMU, Dal, Acadia partnership effort. |

| Other investigators on this project: | Their email addresses: |
|---|---|
| Rinda Digamarthi, MSc student and RA, Acadia | 152527t@acadiau.ca |
| Yigit Aydede, Professor, Saint Mary's University | yigit.aydede@smu.ca |
| Mutlu Yuksel, Professor, Dalhousie University | multu@dal.ca |
| Funders/sponsors of your project (if any): | Nova Scotia COVID-19 Health Research Coalition |
| | |
| Proposed start date of your research: | September 21, 2019     **(4-6 weeks are required for review.)** |

**Enter the date of your application below to certify that you will follow all TCPS2 regulations and REB requirements in conducting your research.**

| |
|---|
| September 14, 2020 |

Lasted Revised:  December 23, 2021

**For student researchers, enter the date below on which your supervisor approved your submission of this application. You must also "cc" your supervisor on your email submission of this application.**

**Research Summary**

1. **Purpose**: the objectives of the study; its hypotheses (if any); why the study is needed

   A collaborative partnership, called the Nova Scotia COVID-19 Health Research Coalition, to develop a COVID-19 research response strategy has formed among NSHA Research & Innovation (Nova Scotia Health Authority), Dalhousie University, Research Nova Scotia, the Dalhousie Medical Research Foundation, the QEII Health Sciences Centre Foundation, the IWK Foundation, and the Dartmouth General Hospital Foundation (DGHF). These partners have collectively committed a minimum $1.5 million to support the Nova Scotia research community. This research effort will inform the best COVID-19 practices and support healthcare decision making and planning that benefits the population of Nova Scotia.

   The proposal submitted by our research team (Yigit Aydede - SMU, Mutlu Yuksel - DAL, Daniel Silver - Acadia) was selected by the Coalition for funding of $36,900 (see the attached application and award notice).  $12,000 from this award will come to Acadia for a Masters level research project in the School of Computer Science.  Using Nova Scotia COVID-19 test data, the proposed study plans to analyze the transmission of viral pathogens and the number of positive COVID-19 cases in response to local climatic and air quality conditions as well as the level and the mode of mobility in Nova Scotia. We will analyze these data with respect to the speed of transmission measured by the demand for the COVID-19 tests and the number of positive tests by region by time. Advance statistical time series methods and deep recurrent neural networks will be use to develop models to predict the spread of the disease.  As input variables, we will use local mobility data extracted from Apple, Google, and Facebook application programming interfaces (API), and high-dimensional high-resolution local weather and air quality data obtained from industry providers such as Breezometer and Climacell.

   While awaiting the data from the Nova Scotia Health Authority (NSHA), the SMU and Acadia team will work with a preliminary dataset from the Province of Ontario that is publicly available by county at Ontario.ca (specifically: https://data.ontario.ca/dataset/confirmed-positive-cases-of-covid-19-in-ontario). Mobility and high resolution weather data will be obtained for these county regions in the same manner as it will be for postal code regions in Nova Scotia.   This will allow us to hone our knowledge of the problem and modeling skills prior to receiving the NS data.

2. **Methodology**: how the subjects will be chosen, how they will be contacted, and by whom; who will conduct the research and where; what the subjects will be asked to do; what data will be taken

   Our analysis combines information from multiple data sources on (i) the nonpharmaceutical interventions measured by social mobility indices of Apple, Google, and Facebook, (ii) the daily number of infections abstracted from 811 Triage, and (iii) high-dimensional high-resolution weather and air quality data.
   The daily data on the number of infections (Covid19 or common flu) at the postal code level will be extracted from health authorities. Since COVID19 tests for the public are regulated by 811 Triage, each referral by 811 Triage represents a significant level of pre-determined symptoms. This provides us with information on the number of new infections each day by postal code. The observed variation in the incidence of viral infections may be influenced in part by variation in reporting. In order to remove such random variation from the data, we may use a moving average of the data to better estimate days with zero or very low cases.

SMU will apply advanced time series statistical modeling methods. They will develop Autoregressive Distributed Lag (ARDL) models with fixed-effect dummies and ARDL with a random-coefficient structure. They have two explicit objectives: prediction and model selection so that sparsity can be achieved by different objective functions. First, SMU wish to determine if there is an underlying data generating a process for the infection transmission that can be captured by a model that represents the "true" sparsity with relevant variables (predictors). They will apply an Adaptive Lasso regression proven to be consistent when applied to time-series and panel data. Second, SMU will find a model that has the most predictive power in forecasting the transmission rate. To accomplish this, they will employ cross-validation methods to find the optimal penalization in a range of Lasso family applications from Elastic Net to Bootstrapped- Lasso. They will also experiment with multiple nonparametric machine learning models specifically suitable for a time series problems.

Acadia will have very similar objectives: (1) To develop the most predictive model possible using a time-series moving window cross-validation approach and (2) To determine the key predictor variables and their potential non-linear interaction to produce accurate estimates of viral disease spread.  This will be accomplished by using state-of-the-art Long Short-Term Memory (LSTM) artificial neural networks, which is a kind of recurrent neural network (RNN) and one of the best predictive models given the chain-like nature of time series and panel data. Acadia's work will focus on predicting the number of new cases of COVID-19 each day, secondarily the Acadia team will look at models that classify the next day as having more or less COVID-19 cases than the day before. Sensitivity analysis will be done on the models to provide insight into the relationship between the input variables and response variable. Less accurate decision tree models may be used to provide some insight into these relationships if it seems warranted.

3. **Consent**: how informed consent will be obtained (**Note: The REB requires parent/guardian consent for any research subject under 18 years of age who is not a registered student at Acadia University**.)

   **All health care data has been collected by provincial health authorities.**
   **The preliminary data that we are using from the Province of Ontario has already been collected and is now in the public domain and freely available to all researchers. The data consists of the number of COVID cases that have tested positive per day per region. The NSHA has already collected the COVID19 test data for Nova Scotia. They are preparing the data for our use with privacy and integrity in mind such that the data cannot be traced back to any individuals. There is no identifying information beyond the region name (county for ON data, postal code for NS data). We will throw away any data that has less than 5 test cases per region.**

   Aggregated smartphone mobility will be obtained for each region from publicly accessible web sources using APIs from Apple, Google, and Facebook.

   The high-resolution weather data will come from industry web sources such as Breezometer or Climacell.

4. **Debriefing**: how the subjects will be debriefed following their participation

    Not applicable. Neither the PI nor RA will be collecting data from subjects.

5. **Risks**: any expected risks to the subjects and how such risks will be minimized

    Not applicable. Neither the PI nor RA will be collecting data from subjects.

    There is no risk because the data (COVID-19 Test numbers) will be aggregated by region and time at NSHA.   The preliminary Ontario data being used is in the public domain and is similarly aggregated by region and time.

6. **Safety**: if applicable, how the safety of subjects will be monitored

    Not applicable. Neither the PI nor RA will be collecting data from subjects.

7. **Confidentiality**: how the confidentiality of the subjects and data will be assured

    **The data consists of the number of COVID cases that have tested positive per day per region (county or postal code). The data aggregation has been completed by the respective health authority. We do not have access to individual records.  The health authority has prepared the data for our us with privacy and integrity in mind such that the data cannot be traced back to any individuals.  There is no identifying information beyond the region name (county for ON data, postal code for NS data).  We will throw away any data that has less than 5 test cases per region.**

    **The data will  be contained in the researchers' local computers.  If requested by NSHA, the data will be transmitted in an encrypted format over the Internet,  and the encryption key will be sent independently by text message.  The Ontario data can be transmitted in open format as it is in the public domain.**

8. **Compensation**: how, if at all, the subjects will be compensated

    Not applicable

9. **Deception**: if deception will be used, why it is necessary

    Not applicable

**Consent forms that will be used**

Not applicable.

**Surveys, questionnaires, or interview questions that will be used**

Not applicable.

**Advertisements that will be used to alert or attract research subjects**

Not applicable.

**Research protocols, if any**

Not applicable.

**Contract agreements, if any**

**Please see the attached NSHA application and award and the Saint Mary's University letter of funds transfer to Acadia University.**

**Confidentiality agreements, if any, between the researcher and his/her source of funding**

**Not applicable.**

B.  Meta Data Report for Original Source Data

The Meta Data Report for all the variables considered during this project can be found at the following [link](link).

C.  Meta Data Report for Prepared Modeling Data

The Meta Data Report for just those variables used in the development of machine learning models can be found at the following [link](link).